\documentclass{article} 
\usepackage[preprint]{acl}
\usepackage{times}

\usepackage{amsmath,amsfonts,bm}

\def\eqref#1{equation~\ref{#1}}

\def\1{\bm{1}}

\DeclareMathAlphabet{\mathsfit}{\encodingdefault}{\sfdefault}{m}{sl}
\SetMathAlphabet{\mathsfit}{bold}{\encodingdefault}{\sfdefault}{bx}{n}

\usepackage{hyperref}
\usepackage{float}
\usepackage{url}
\usepackage{graphicx}
\usepackage{booktabs}   
\usepackage{multirow}
\usepackage{pdflscape}
\usepackage{booktabs}
\usepackage{multirow}
\usepackage{caption}     
\usepackage{pgfplots}
\pgfplotsset{compat=1.18}
\usepackage[table,xcdraw]{xcolor}
\usepackage{booktabs}     
\usepackage{subcaption}   
\usepackage{multirow}     
\usepackage{adjustbox}    
\usetikzlibrary{patterns}
\usepackage{sansmath} 
\usepackage{amssymb}
\usepackage{utfsym}
\usepackage{amsthm}
\usepackage{algorithm}
\usepackage{listings}
\usepackage{minted}
\usepackage[section]{placeins}
\usepackage[T1]{fontenc}

\theoremstyle{definition}
\newtheorem{definition}{Definition}

\newcommand{\benchname}{Needle in the Web}

\title{\benchname: A Benchmark for Retrieving Targeted Web Pages in the Wild}

\author{Yumeng Wang \\ Tsinghua University \\ \texttt{wangyume22@mails.tsinghua.edu.cn}
        \AND 
        Tianyu Fan \and Lingrui Xu \and Chao Huang\thanks{Chao Huang is the Corresponding Author.} \\ The University of Hong Kong \\ \texttt{\{tianyufan0504, lingruixu.db, chaohuang75\}@gmail.com}}

\begin{document}

\maketitle

\begin{abstract}
Large Language Models (LLMs) have evolved from simple chatbots into sophisticated agents capable of automating complex real-world tasks, where browsing and reasoning over live web content is key to assessing retrieval and cognitive skills. Existing benchmarks like BrowseComp and xBench-DeepSearch emphasize complex reasoning searches requiring multi-hop synthesis but neglect Fuzzy Exploratory Search, namely queries that are vague and multifaceted, where users seek the most relevant webpage rather than a single factual answer. To address this gap, we introduce ~\textbf{\benchname}, a novel benchmark specifically designed to evaluate modern search agents and LLM-based systems on their ability to retrieve and reason over real-world web content in response to ambiguous, exploratory queries under varying levels of difficulty. \benchname~comprises 663 questions spanning seven distinct domains. To ensure high query quality and answer uniqueness, we employ a flexible methodology that reliably generates queries of controllable difficulty based on factual claims of web contents. We benchmark three leading LLMs and three agent-based search systems on \benchname, finding that most models struggle: many achieve below 35\% accuracy, and none consistently excel across domains or difficulty levels. These findings reveal that \benchname presents a significant challenge for current search systems and highlights the open problem of effective fuzzy retrieval under semantic ambiguity. Data and code are available at \hyperlink{https://github.com/Tango-Whiskyman/Needle\_in\_the\_Web}{https://github.com/Tango-Whiskyman/Needle\_in\_the\_Web}.
\end{abstract}

\section{Introduction}

\pgfplotsset{compat=1.17}

\definecolor{ModelA}{RGB}{33,158,188}   
\definecolor{ModelB}{RGB}{255,145,0}    
\definecolor{ModelC}{RGB}{142,68,173}   
\definecolor{ModelD}{RGB}{38,60,83}     
\definecolor{ModelE}{RGB}{220,20,60}    
\definecolor{ModelF}{RGB}{34,139,34}    

\begin{figure*}[h!]
\centering
\begin{tikzpicture}

\begin{axis}[
  every axis plot/.append style={forget plot},
  ybar,
  width=\linewidth, height=8.4cm,
  ymin=0, ymax=60,
  xmin=-0.5, xmax=47.5,
  axis lines=none,
  xtick={2.5,9.5,16.5,23.5,30.5,37.5,44.5},
  tick label style={font=\scriptsize},
  ymajorgrids, grid style={black!10},
  bar width=4.5pt,
  clip=false,
  enlarge x limits=0.01,
  every axis/.append style={line width=0.4pt, tick style={line width=0.4pt}},
]
\addplot+[draw=ModelA!60, fill=ModelA] coordinates {
  (0,30.81) (7,41.28) (14,51.53) (21,16.20) (28,38.10) (35,37.16) (42,14.50)
};
\end{axis}

\begin{axis}[
  every axis plot/.append style={forget plot},
  ybar,
  width=\linewidth, height=8.4cm,
  ymin=0, ymax=60,
  xmin=-0.5, xmax=47.5,
  axis lines=none,
  xtick={2.5,9.5,16.5,23.5,30.5,37.5,44.5},
  tick label style={font=\scriptsize},
  ymajorgrids, grid style={black!10},
  bar width=4.5pt,
  clip=false,
  enlarge x limits=0.01,
  every axis/.append style={line width=0.4pt, tick style={line width=0.4pt}},
]
\addplot+[draw=ModelB!60, fill=ModelB] coordinates {
  (1,27.40) (8,32.36) (15,44.74) (22,22.73) (29,40.89) (36,19.09) (43,22.56)
};
\end{axis}

\begin{axis}[
  every axis plot/.append style={forget plot},
  ybar,
  width=\linewidth, height=8.4cm,
  ymin=0, ymax=60,
  xmin=-0.5, xmax=47.5,
  axis lines=none,
  xtick={2.5,9.5,16.5,23.5,30.5,37.5,44.5},
  tick label style={font=\scriptsize},
  ymajorgrids, grid style={black!10},
  bar width=4.5pt,
  clip=false,
  enlarge x limits=0.01,
  every axis/.append style={line width=0.4pt, tick style={line width=0.4pt}},
]
\addplot+[draw=ModelC!60, fill=ModelC] coordinates {
  (2,47.60) (9,48.74) (16,42.76) (23,25.03) (30,42.00) (37,19.09) (44,12.41)
};
\end{axis}

\begin{axis}[
  every axis plot/.append style={forget plot},
  ybar,
  width=\linewidth, height=8.4cm,
  ymin=0, ymax=60,
  xmin=-0.5, xmax=47.5,
  axis lines=none,
  xtick={2.5,9.5,16.5,23.5,30.5,37.5,44.5},
  tick label style={font=\scriptsize},
  ymajorgrids, grid style={black!10},
  bar width=4.5pt,
  clip=false,
  enlarge x limits=0.01,
  every axis/.append style={line width=0.4pt, tick style={line width=0.4pt}},
]
\addplot+[draw=ModelD!60, fill=ModelD] coordinates {
  (3,10.48) (10,45.45) (17,44.98) (24,24.95) (31,33.91) (38,36.75) (45,17.35)
};
\end{axis}

\begin{axis}[
  every axis plot/.append style={forget plot},
  ybar,
  width=\linewidth, height=8.4cm,
  ymin=0, ymax=60,
  xmin=-0.5, xmax=47.5,
  axis lines=none,
  xtick={2.5,9.5,16.5,23.5,30.5,37.5,44.5},
  tick label style={font=\scriptsize},
  ymajorgrids, grid style={black!10},
  bar width=4.5pt,
  clip=false,
  enlarge x limits=0.01,
  every axis/.append style={line width=0.4pt, tick style={line width=0.4pt}},
]
\addplot+[draw=ModelE!60, fill=ModelE] coordinates {
  (4,41.54) (11,44.44) (18,42.98) (25,20.60) (32,30.78) (39,33.87) (46,14.40)
};
\end{axis}

\begin{axis}[
  every axis plot/.append style={forget plot},
  ybar,
  width=\linewidth, height=8.4cm,
  ymin=0, ymax=60,
  xmin=-0.5, xmax=47.5,
  ylabel={\small Accuracy (\%)},
  axis y line*=left, axis x line*=bottom,
  xtick={2.5,9.5,16.5,23.5,30.5,37.5,44.5},
  xticklabels={ArXiv, OLH, Wikipedia, CNN, Petapixel, Pitchfork, Lonelyplanet},
  xticklabel style={font=\scriptsize, align=center},
  tick label style={font=\scriptsize},
  ymajorgrids, grid style={black!10},
  bar width=4.5pt,
  legend style={draw=none, fill=none, font=\scriptsize, row sep=1pt},
  legend pos=north east,
  clip=false,
  enlarge x limits=0.01,
  every axis/.append style={line width=0.4pt, tick style={line width=0.4pt}},
]
\addplot+[draw=ModelF!60, fill=ModelF] coordinates {
  (5,12.88) (12,16.43) (19,13.33) (26,20.60) (33,14.75) (40,6.19) (47,2.95)
};

\addlegendimage{area legend, draw=none, fill=ModelA}
\addlegendentry{GPT-4o}

\addlegendimage{area legend, draw=none, fill=ModelB}
\addlegendentry{Gemini-2.5-flash}

\addlegendimage{area legend, draw=none, fill=ModelC}
\addlegendentry{Perplexity Sonar}

\addlegendimage{area legend, draw=none, fill=ModelD}
\addlegendentry{Search-R1}

\addlegendimage{area legend, draw=none, fill=ModelE}
\addlegendentry{DeepResearcher}

\addlegendimage{area legend, draw=none, fill=ModelF}
\addlegendentry{CognitiveKernel-Pro}

\draw[black!30, densely dashed] (axis cs:6,0) -- (axis cs:6,60);
\draw[black!30, densely dashed] (axis cs:13,0) -- (axis cs:13,60);
\draw[black!30, densely dashed] (axis cs:20,0) -- (axis cs:20,60);
\draw[black!30, densely dashed] (axis cs:27,0) -- (axis cs:27,60);
\draw[black!30, densely dashed] (axis cs:34,0) -- (axis cs:34,60);
\draw[black!30, densely dashed] (axis cs:41,0) -- (axis cs:41,60);

\end{axis}

\end{tikzpicture}
\caption{An overview of model performance on \benchname. Items on X-axis denote the source websites from which queries are collected.}
\label{fig:singlebars-iclr}
\end{figure*}
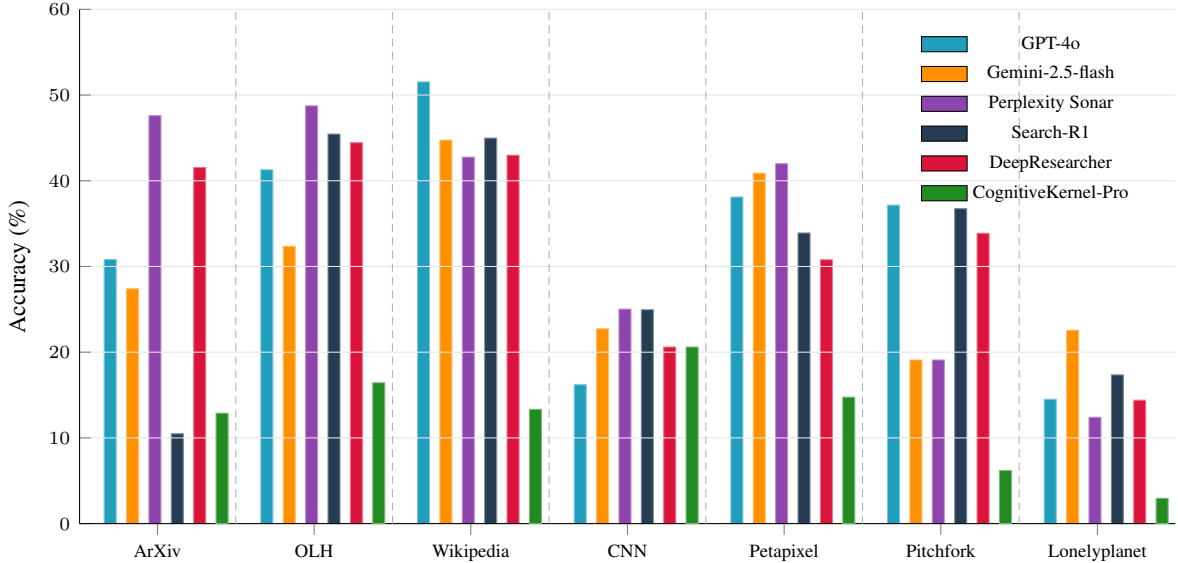


Large Language Models (LLMs) are rapidly evolving from simple chatbots into more agentic systems capable of autonomously invoking tools and making decisions~\citep{chowa_language_2025,tang_autoagent_2025}. Given the complexity and dynamism of real-world environments, the ability to search—i.e., to retrieve information from the internet—has become an essential component in building effective agents, and navigating the vast expanse of online information now serves as a critical test of an agent's capabilities~\citep{google_gemini_2025,openai_introducing_2025,perplexityai_introducing_2025,liu_webglm_2023,he_webvoyager_2024}


To measure agents’ capabilities in web navigation and online content comprehension, researchers have proposed several evaluation benchmarks \citep{chen_browsecomp-plus_2025,zhou_browsecomp-zh_2025,zhou_academicbrowse_2025,tao_mmsearch-plus_2025,trivedi__2022,yang_hotpotqa_2018}.  
Among these, the most representative are BrowseComp \citep{wei_browsecomp_2025} and xBench-DeepSearch \citep{xbench-team_xbench-deepsearch_2025}. A common characteristic of these benchmarks is \textit{Complex Reasoning Search}, i.e., queries in this paradigm are typically explicit and highly structured, often relying on multi-hop queries with key information masked, ultimately requiring a single correct answer. For instance, tasks in BrowseComp are deliberately designed to include numerous direct constraints (e.g., detailed descriptions of character attributes, timelines, and contextual details), making it impossible to derive the answer through a single direct lookup. A successful agent must continuously navigate across multiple webpages, connect scattered pieces of evidence, and finally produce a concise, verifiable answer that is fully supported by all the collected clues.

\begin{figure*}[h]
\begin{center}
\includegraphics[width=\linewidth]{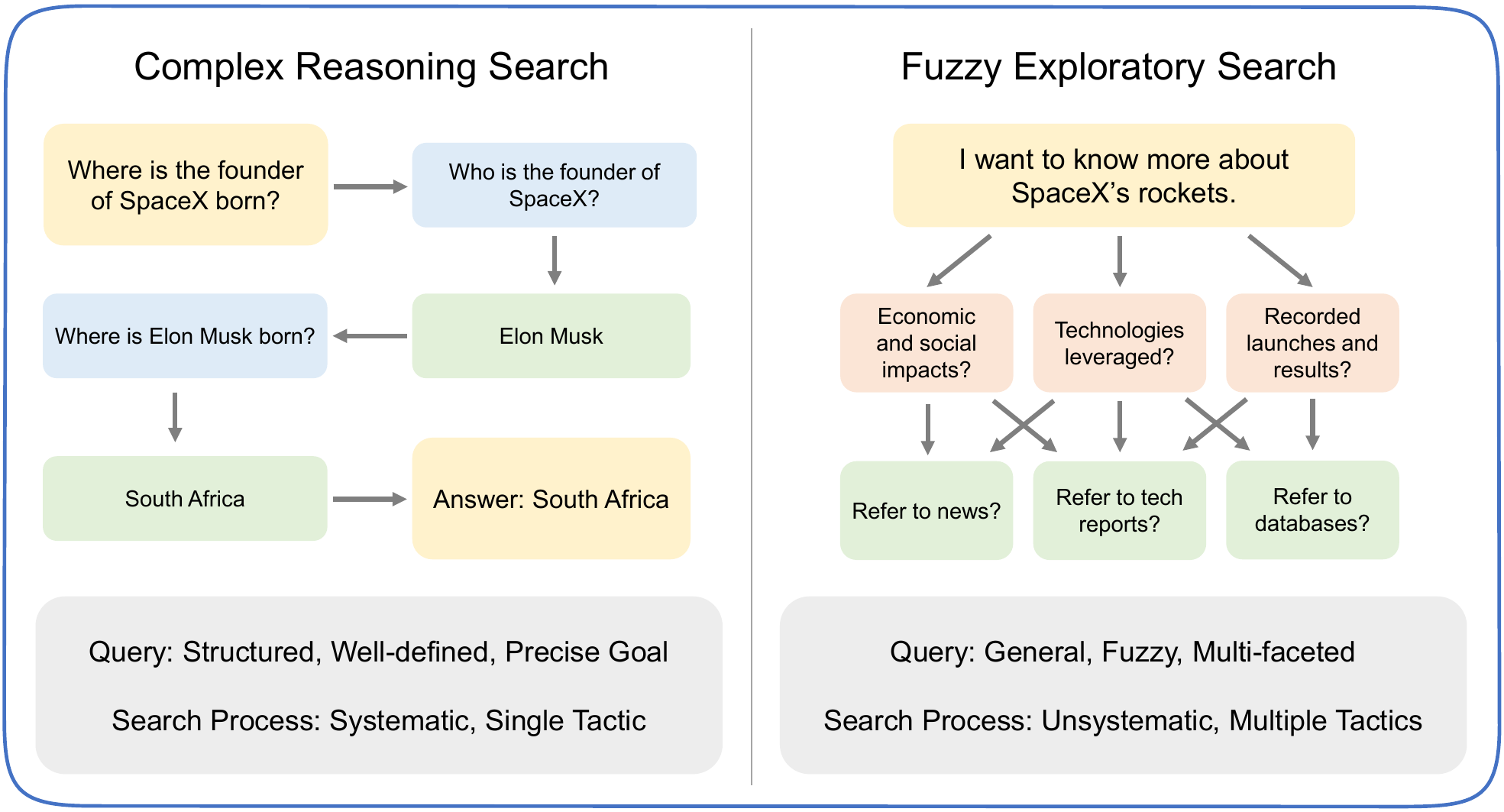}
\end{center}
\caption{A comparison between Complex Reasoning Search and Fuzzy Exploratory Search. Complex Reasoning Search follows a clear strategy and only involves factoid information. Fuzzy Exploratory Search, on the contrary, must deal with multi-faceted queries. It needs to identify the query's implicit requirements and find the most appropriate source.}
\label{fig_search_paradigms}
\end{figure*}

Although the \textit{Complex Reasoning Search} scenario has been thoroughly explored, the more realistic user-oriented \textit{Fuzzy Exploratory Search} has received comparatively little attention. In real-world environments, users often issue queries that are vague, multi-dimensional, or semantically ambiguous, with their search intent not expressed as a precise question \citep{liu_mining_2024}. In such cases, the goal is not merely to retrieve a short factual answer, but rather to identify information from appropriate sources that best aligns with the user’s implicit criteria. As illustrated in Figure~\ref{fig_search_paradigms}, a user might issue an instruction such as: \textit{I’d like to learn more about SpaceX’s rockets}. As the search progresses and results are examined, the user may further refine or elaborate their requirements regarding the content of retrieved webpages. For such broad and open-ended queries, there is no single, obvious correct answer. However, by recognizing the underlying dimensions of the query and the user’s latent information needs, the most appropriate response would be specific webpages or articles that best satisfy those needs.
Given the open-ended, iterative, and multi-faceted nature of \textit{Fuzzy Exploratory Search} queries \citep{soufan_searching_2022, medlar_unexplored_2024}, designing evaluable queries becomes considerably challenging. In \textit{Fuzzy Exploratory Search}, users often refine and deepen their understanding after seeing initial results, typically seeking a relevant passage or a set of resources rather than a concise, singular answer.

To address this gap, we introduce \textbf{\benchname~(NiW)}—a novel benchmark specifically designed to evaluate the performance of search agents when handling fuzzy, exploratory web queries, spanning multiple levels of difficulty. Unlike traditional factoid QA, this benchmark requires agents to find a needle in a vast haystack: precisely identifying the single webpage that best matches a given set of ambiguous and underspecified criteria from among countless candidates on the open web.
Our benchmark comprises 663 queries spanning seven diverse domains, from computer science and humanities to travel blogs and everyday news, carefully constructed to simulate the kind of fuzzy query a real user might pose. We devise a flexible query generation methodology that reliably produces queries of controllable difficulty based on factual web content. Further evaluation confirms that our method reliably limits the number of webpages that match all the implicit criteria to its minimum, excluding superficial or partial matches. By adjusting the amount of fuzziness, we can dial the difficulty of each query, enabling fine-grained evaluation of an agent's retrieval skill.

We benchmark a representative set of state-of-the-art LLM-based search agents on \benchname, including three leading closed-source models and three open-source frameworks. The evaluation reveals that fuzzy exploratory search poses a formidable challenge to current agents. Despite their impressive capabilities in standard Q\&A and even multi-hop reasoning, most agents struggle on our benchmark's queries. In our evaluation, a large portion of agents achieve below 35\% accuracy, with only a few exceptions. Notably, no single agent dominates across all domains or difficulty levels. This inconsistency highlights the substantial gap between today's LLM retrieval capabilities and the requirements of truly robust web search. Moreover, no agent consistently and significantly outperformed all others across our evaluations.
Our experiments demonstrate that, for LLMs, accurately locating a web page that achieves deep semantic alignment with a fuzzy query remains an unsolved challenge. This highlights that \benchname can serve as a diagnostic benchmark to drive technical progress in the field and incentivize the development of more advanced LLM-powered search agents capable of effectively handling the complexities of exploratory search in real-world web environments.


\section{Related work}

\textbf{Search Agents.}
To equip agents with access to external knowledge, early research commonly employed Retrieval-Augmented Generation (RAG) to dynamically inject knowledge from external data sources into the LLM’s prompt~\citep{guo_lightrag_2025,xie_weknow-rag_2024,fan_minirag_2025}, thereby mitigating hallucinations and improving factual accuracy \citep{lewis_retrieval-augmented_2020, gao_retrieval-augmented_2023}. Building on this paradigm, some approaches train search agents using supervised fine-tuning (SFT). However, SFT alone severely limits the agent’s generalization capability.
Recently, owing to the strong generalization exhibited by reinforcement learning (RL), researchers have begun exploring RL-guided agent architectures, shifting the training of search agents toward an RL-based paradigm~\citep{jin_search-r1_2025, zheng_deepresearcher_2025, fang_cognitive_2025}. These RL-based search agents demonstrate more advanced capabilities, including multi-step web navigation, information synthesis, and iterative query refinement.


\textbf{Benchmarks for Search Agents.}
Evaluation of search agents is typically conducted using question–answer (Q–A) pairs: given a query, the agent autonomously searches for relevant information and its response is judged for correctness. As is shown in Fig.\ref{tab:benchmark_comparison}, there are several principles crucial to holistic evaluation, in which existing QA benchmarks for search agents differ: \textbf{(1) Live Web Retrieval.} An ideal benchmark should require agents to search the open web and to interact with webpages. While recent works fulfill this requirement, early QA tasks were largely confined to closed corpora, including benchmarks such as HotpotQA~\citep{yang_hotpotqa_2018}, 2WikiMultiHopQA~\citep{ho_constructing_2020}, and MuSiQue~\citep{trivedi__2022}. \textbf{(2) Difficulty Control.} Dividing queries into distinct difficulties enables a more fine-grained evaluation. Current criteria of determining difficulty include the length of reasoning chain, i.e. the number of hops in multi-hop queries~\citep{trivedi__2022}, and the amount of missing information that must be searched for~\citep{tao_mmsearch-plus_2025}. \textbf{(3) Multi-Scenario Coverage.} As search activities of real users span multiple topics and websites, benchmarks should represent as many of them as possible. Older widely used benchmarks, HotpotQA, 2WikiMultiHopQA, and MuSiQue, focus solely on Wikipedia and thus fail to address other user scenarios. In contrast, recent benchmarks, unless designed for a specific purpose like academic search~\citep{zhou_academicbrowse_2025}, fulfill this requirement. \textbf{(4) Full Webpage Retrieval and (5) Fuzzy Exploratory Query.} Users often pose Fuzzy Exploratory Search tasks without a closed-source ground truth. For this type of query, finding appropriate source of information is essential. Existing benchmarks overlook these aspects as they follow the Complex Reasoning Search paradigm, persistently deepening multi-hop reasoning and focusing exclusively on queries with specific factoid answers~\citep{xbench-team_xbench-deepsearch_2025, zhou_browsecomp-zh_2025, chen_browsecomp-plus_2025, zhou_academicbrowse_2025, tao_mmsearch-plus_2025}.

\section{\benchname}

In this section, we demonstrate the \benchname~pipelines that we developed. These pipelines enable us to automatically and reliably collect queries and evaluate agent responses.

\begin{figure*}[h]
\begin{center}
\includegraphics[width=\linewidth]{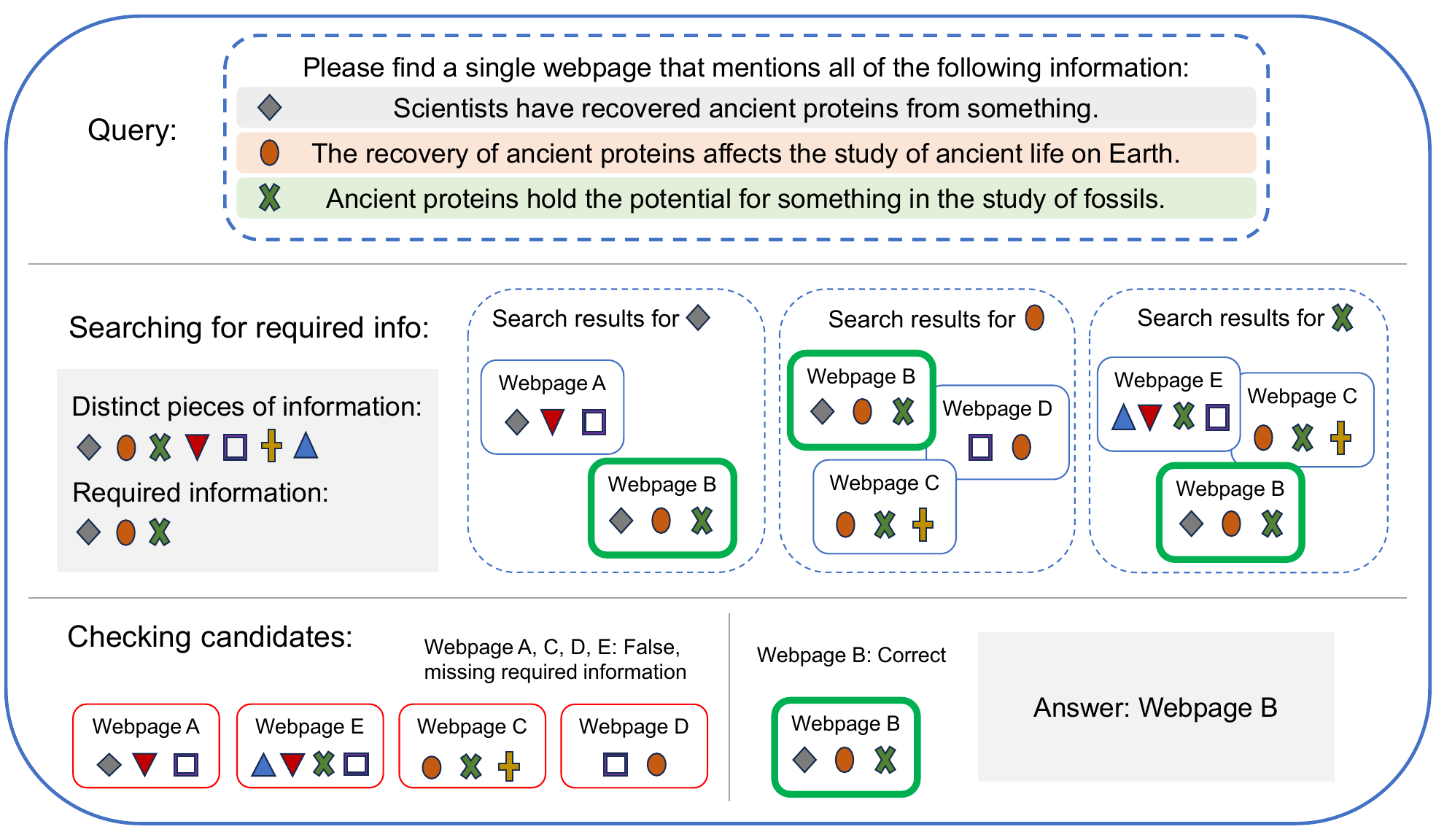}
\end{center}
\caption{A sample query of \benchname. Each of the separate requirements may be satisfied by multiple webpages, yet only the webpage that meets all requirements is considered the correct answer.}
\label{fig_query_example}
\end{figure*}

\subsection{Incorporating vague queries by query design}
\label{sec:design}

\begin{table*}[t!]
\centering
\renewcommand{\arraystretch}{1.25}
\setlength{\tabcolsep}{8pt}
\resizebox{\textwidth}{!}{
\begin{tabular}{lcccccc}
\toprule
\multirow{2}{*}{\textbf{Benchmark}} &
\textbf{Live Web} &
\textbf{Difficulty-} &
\textbf{Multi-Scenario} &
\textbf{Full Webpage} &
\textbf{Fuzzy Exploratory} &
\\
&
\textbf{Retrieval} &
\textbf{Controlled} &
\textbf{Coverage} &
\textbf{Retrieval} &
\textbf{Query} &
\\
\midrule
HotpotQA \citep{yang_hotpotqa_2018} & ${\color{red}\usym{2717}}$ & ${\color{red}\usym{2717}}$ & ${\color{red}\usym{2717}}$ & ${\color{red}\usym{2717}}$ & ${\color{red}\usym{2717}}$  \\
MuSiQue \citep{trivedi__2022} & ${\color{red}\usym{2717}}$ & ${\color{green}\usym{2713}}$ & ${\color{red}\usym{2717}}$ & ${\color{red}\usym{2717}}$ & ${\color{red}\usym{2717}}$  \\
AcademicBrowse \citep{zhou_academicbrowse_2025} & ${\color{green}\usym{2713}}$ & ${\color{red}\usym{2717}}$ & ${\color{red}\usym{2717}}$ & ${\color{red}\usym{2717}}$ & ${\color{red}\usym{2717}}$  \\
BrowseComp \citep{wei_browsecomp_2025} & ${\color{green}\usym{2713}}$ & ${\color{red}\usym{2717}}$ & ${\color{green}\usym{2713}}$ & ${\color{red}\usym{2717}}$ & ${\color{red}\usym{2717}}$  \\
XBench-DeepSearch \citep{xbench-team_xbench-deepsearch_2025} & ${\color{green}\usym{2713}}$ & ${\color{red}\usym{2717}}$ & ${\color{green}\usym{2713}}$ & ${\color{red}\usym{2717}}$ & ${\color{red}\usym{2717}}$ &  \\
MMSearch-Plus \citep{tao_mmsearch-plus_2025} & ${\color{green}\usym{2713}}$ & ${\color{green}\usym{2713}}$ & ${\color{green}\usym{2713}}$ & ${\color{red}\usym{2717}}$ & ${\color{red}\usym{2717}}$  \\

\midrule
\textsc{\benchname~(Ours)} & ${\color{green}\usym{2713}}$ & ${\color{green}\usym{2713}}$ & ${\color{green}\usym{2713}}$ & ${\color{green}\usym{2713}}$ & ${\color{green}\usym{2713}}$  \\
\bottomrule
\end{tabular}
}
\caption{Comparison of \benchname~with existing benchmarks for search agents.}
\label{tab:benchmark_comparison}
\end{table*}


In real-world information retrieval scenarios, a large proportion of user queries are inherently ambiguous, often stemming from incomplete, imprecise, or even only implicitly expressed intents. Fig.~\ref{fig_query_example} showcases an example of the queries of \benchname. It is inspired by how humans retrieve information based only on vague, implicit requirements. The example query could represent a typical scenario in which a person tries to remind themselves of an article titled \textit{Scientists recover proteins from a 24 million-year-old rhino fossil} that they have skimmed through earlier. However, due to the vagueness of human memory, the user typically cannot accurately recall the article’s title, author, or other explicit metadata, and instead retains only a few semantically related but broadly phrased impressions, such as \textit{the article mentioned scientists successfully extracting proteins from some ancient material.}

Such retrieval tasks can be formalized as a multi-constraint information retrieval problem, wherein each vague impression acts as an implicit semantic constraint on the target document. The core design objective of the~\benchname~dataset is precisely to evaluate an agent’s ability to effectively integrate and reason over multiple implicit semantic cues under such ambiguous, multi-constraint conditions. Notably, any single impression often corresponds to a vast set of potential documents, yet the number of documents satisfying all implicit constraints simultaneously is typically extremely small—even unique, as shown in later evaluations. It is precisely this intersection property of constraints that enables human users, despite lacking precise keywords, to efficiently and reliably locate target content in open-web environments. This characteristic also provides a critical evaluation dimension for developing agents with human-like retrieval capabilities.

The queries in our benchmark represent this kind of fuzzy exploratory search tasks, which are vague and multifaceted, and do not express the search intent as a precise factoid query. Specifically, the queries in our benchmark provide the agent several pieces of vague information where a central part is masked using a generic expression (e.g. replacing a person's name with \textit{someone}), and require the agent to find a specific webpage whose content mentions all of the information provided. We do not ask the agent to find the masked elements and specify the vague information. Instead, we demand that the agent find exactly one webpage that contains all the provided information. We adopt this setting, because we do not want the agent to compose information from different sources. First, each piece of information required by a query may be mentioned in distinct webpages. Since these pages are not necessarily \textit{talking about the same thing}, simply assembling these pages together as the completion of the vague information is meaningless. Second, false or inaccurate information exists on the internet. Regarding a same event, one may find in the internet distinct articles with conflicting statements. Naturally, for the agents to return false information is undesirable. 
Given that accurately assessing the veracity of information in open-web environments remains a highly challenging task for current agents, even those equipped with real-time web access~\citep{yao_fact-checking_2025, barkett_reasoning_2025}, we do not treat fact-checking or information verification as part of our evaluation objective. Instead, we adopt a more pragmatic and operational constraint: agents should not construct answers based on contradictory or fragmented information from disparate sources, as such responses themselves constitute clear errors.  
Inspired by this principle, our query design is grounded in a key observation: information within a single webpage is typically internally consistent and free of logical contradictions. Therefore, by requiring that the complete answer must appear verbatim within a single webpage, we effectively avoid the semantic inconsistencies and factual risks associated with cross-source synthesis. This approach provides a reliable and controllable benchmark for evaluating agents’ precise retrieval capabilities under ambiguous, multi-constraint conditions.

Overall, our queries measure two core abilities that were overlooked in previous benchmarks: 1) The ability of interacting well with the search tool, including reasonably identifying the optimal search keywords from a vague query, and correctly representing the search results. 2) The ability of carefully examining the results obtained to identify the correct webpage to return.

\subsection{Automated query collection}

\begin{figure*}[h]
\begin{center}
\includegraphics[width=\linewidth]{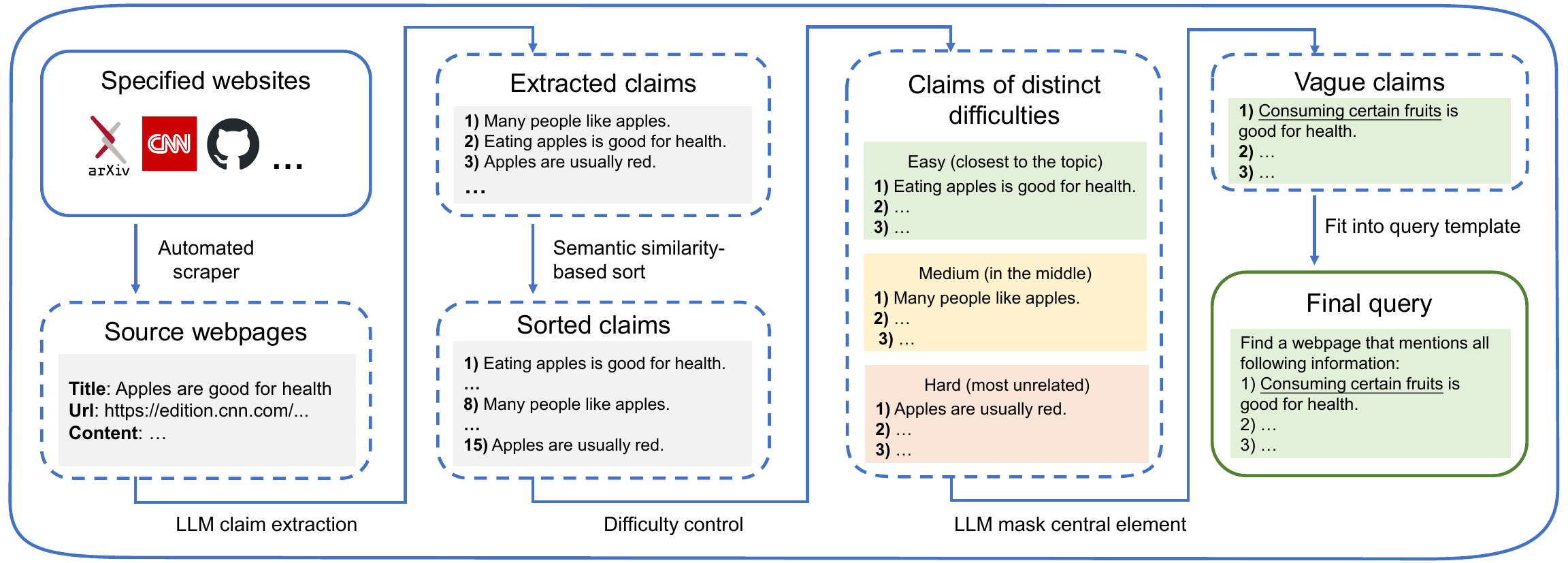}
\end{center}
\caption{An illustration of our automated query collection pipeline. Different selected claims undergo the same processing, their only difference is in the difficulty of final query.}
\label{fig_query_collection}
\end{figure*}

To construct a large number of high-quality queries in a consistent and scalable manner, we develop an automated pipeline that transforms real web articles into fuzzy exploratory queries. After selecting several websites that represent different domains, we use FireCrawl~\citep{firecrawl_firecrawl_2024} to automatically scrape article content from each domain and use these articles as the basis for query generation. 

Our pipeline begins by extracting factual claims from an article. Let $\mathcal{D}$ denote the set of all articles. For each article $d$, we prompt an LLM to identify and rewrite involved factual statements in the form of short declarative sentences. We use $C(d)=\{c_1,\dots,c_{m_d}\}$ to denote the set of factual claims extracted from $d$, where each $c_i$ is a short declarative proposition. Next, to estimate each claim’s thematic relevance to the article as a whole, we compute embeddings for both the full article $d$ and all extracted claims $C(d)$ using OpenAI's text-embedding-3-large model~\citep{openai_text-embedding-3-large_2024}. We then rank the claims by their semantic similarity to the article: the most central claims appear at the top of the list, while increasingly tangential claims appear later.

This ranking enables fine-grained control over query difficulty. We assume that a query becomes harder when it is composed of claims that are less relevant to the article's main theme. Accordingly, we categorize difficulties by selecting different segments of the ranked list:

\begin{description}
    \item [Easy] queries use the top three most relevant claims;
    \item [Medium] queries use three mid-ranked claims; and
    \item [Hard] queries use the three least relevant claims.
\end{description}

The claims chosen are denoted by the \textbf{ground truth} of the query. After specifying them, we convert them into vague, non-specific criteria that resemble realistic exploratory search inputs, formally defined as follows:

\begin{definition}[Masked Criterion]
Given a factual claim $c \in C(d)$, a \emph{masked criterion} is obtained by applying a masking function
\[
f_{\mathrm{mask}} : C(d) \to \widetilde{C}(d),
\]
which replaces entity-specific content (names, locations, species, etc.) with generic placeholders such as \textit{someone}, \textit{somewhere}, or \textit{a certain species}.  
We denote the resulting masked predicate by $\tilde c = f_{\mathrm{mask}}(c)$.
\end{definition}

To do so, we prompt an LLM to mask each claim's key entities with generic expressions. The rewritten statements retain their semantic content but omit identifying details. We refer to these rewritten statements as \textbf{criteria}, and they constitute the actual query shown to the agent. For each generated query, we store the original article content and URL, the selected ground-truth claims, and the vague criteria used to form the final question. During evaluation, the criteria are inserted into a fixed query template that instructs the agent to find a webpage mentioning all parts of the criteria.

 To rule out erroneous queries, we immediately validate the queries after generation. For each query, we provide the original article content and the criteria to an LLM, and ask whether the article does mention all parts of the criteria. If any piece of information is judged unsupported, we discard the query. This validation step filters out errors arising from imperfect claim extraction or rewriting.

\subsection{Automated evaluation}



Evaluating the answers requires a specifically tailored approach. Traditional exact-matching scoring is not suitable because a single article may appear under multiple URLs, or be syndicated across different sites. Instead, we draw on prior work demonstrating that LLMs can reliably serve as evaluators~\citep{kamalloo_evaluating_2023, yang_crag-comprehensive_2024, xu_critical_2023}. Importantly, while locating the correct webpage is challenging for agents, verifying whether a specific page satisfies the query criteria is easy and aligns well with the reading comprehension strengths of current LLMs. Our evaluation pipeline therefore adopts an LLM-as-a-judge mechanism.


To determine whether a webpage contains the information expressed by a criterion, we adopt the following notions of \textbf{semantic mention} and \textbf{query satisfaction}:

\begin{definition}[Semantic Mention]
Let $d$ be a document with extracted claims $C(d)=\{c_1,\dots,c_m\}$, and let 
$t$ be a (possibly masked) query criterion.  
We say that $d$ \emph{mentions} $t$ if and only if there exists a claim 
$c \in C(d)$ such that $c$ \emph{textually entails} $t$. Following~\citet{dagan_recognizing_2022}'s definition, a hypothesis $t$ is considered entailed by a text 
$c$ if a competent human reader would typically judge that $t$ is most likely true 
given the information expressed in $c$.  
Our notion of \emph{semantic mention} adopts this definition: a document
mentions a query criterion when at least one of its extracted claims entails
the criterion in this sense.
Formally,
\[
d \models t 
\quad\Longleftrightarrow\quad
\exists\, c \in C(d):\; c \Rightarrow t,
\]
where $c \Rightarrow t$ holds if humans reading $c$ would typically infer 
that $t$ is most likely true, without requiring information beyond common 
background knowledge.
\end{definition}

\begin{definition}[Query Satisfaction]
A document $d$ satisfies a query $q = \{\tilde c_1, \tilde c_2, \tilde c_3\}$ if
\[
d \models \tilde c_1 \ \wedge\ d \models \tilde c_2 \ \wedge\ d \models \tilde c_3.
\]
\end{definition}

For each model-generated answer, we retrieve the webpage indicated by the answer, and extract its main textual content. Subsequently, we supply the query's criteria and extracted answer webpage content to the LLM judge. The judge first determines whether the answer webpage content mentions all criteria. If any criterion is missing, the answer is marked incorrect. If all criteria are present, the judge then checks whether all ground-truth claims are mentioned. Should they be mentioned, the answer is labeled a ground-truth match. Otherwise, it is marked a criteria match. Both cases are considered correct, though only the former corresponds exactly to the expected target page.


\subsection{Benchmark Composition}


To ensure that the benchmark constructed in this study authentically reflects real-world information retrieval demands, we sourced our corpus from repositories that provide large-scale, well-structured articles of moderate length. Guided by this criterion, we systematically collected query samples from seven representative websites spanning both academic research and everyday life domains. The selected websites are: Arxiv Computer Science Repository, Open Library of Humanities (a website for publishing preprints of Humanities), Wikipedia, CNN News, Lonelyplanet travel blogs, Pitchfork (a website of album reviews), and Petapixel (a website featuring digital product reviews). From each website, we randomly sampled 30 to 35 articles as the foundational corpus. Based on each sampled article, we generated one easy, one medium, and one hard query. After rigorously filtering the initially generated queries—removing those suffering from semantic ambiguity, factual inaccuracies, or unverifiability, we ultimately constructed a high-quality evaluation set comprising 663 queries: 222 easy, 229 medium, and 212 hard. This dataset exhibits a well-balanced distribution across query difficulty levels and domain coverage, thereby enabling robust, multi-dimensional evaluation of retrieval system performance.

\section{Experiments}

\newcolumntype{C}[1]{>{\centering\arraybackslash}m{#1}}

\begin{table*}[h]
\centering
\small
\begin{tabular}{l c C{1.6cm} C{1.6cm} C{1.6cm}}
\toprule
\multirow{2}{*}[-0.3em]{\textbf{Model}} &
\multirow{2}{*}[-0.3em]{\textbf{Overall}} &
\multicolumn{3}{c}{\textbf{Accuracy under different difficulties (\%)}} \\
\cmidrule(lr){3-5}
& & \textbf{Easy} & \textbf{Medium} & \textbf{Hard} \\
\midrule
GPT-4o~\citep{hurst_gpt-4o_2024} & 32.88 & \textbf{58.56} & 27.07 & 12.26 \\
Gemini 2.5-flash~\citep{google_gemini_2025} & 30.17 & 46.40 & 30.13 & 13.21 \\
Perplexity Sonar~\citep{perplexityai_models_2025} & \textbf{33.18} & 53.60 & \textbf{31.44} & \textbf{13.68} \\
Search-R1~\citep{jin_search-r1_2025} & 30.77 & 50.90 & 30.57 & 9.91 \\
DeepResearcher~\citep{zheng_deepresearcher_2025} & 32.88 & 57.66 & 27.51 & 12.74 \\
CognitiveKernel-Pro~\citep{fang_cognitive_2025} & 12.37 & 16.67 & 12.66 & 7.55 \\
\bottomrule
\end{tabular}
\caption{An overview of accuracies achieved by different models.}
\label{tab:accuracy-overview}
\end{table*}

\begin{table*}[t]
\centering
\small
\setlength{\tabcolsep}{4pt}
\begin{tabular}{lcccccc}
\toprule
\textbf{Queryset (Difficulty)} & \textbf{GPT-4o} & \textbf{Gemini 2.5-flash} & \textbf{Sonar} & \textbf{Search-R1} & \textbf{DR} & \textbf{CKP} \\
\midrule
ArXiv (Easy)      & 51.52 & 39.39 & \textbf{75.76} & 15.15 & 66.67 & 15.15 \\
ArXiv (Medium)    & 24.24 & 30.30 & \textbf{54.55} & 12.12 & 45.45 & 15.15 \\
ArXiv (Hard)      & \textbf{16.67} & 12.50 & 12.50 & 4.17 & 12.50 & 8.33 \\
\midrule
OLH (Easy)        & 73.53 & 50.00 & 73.53 & \textbf{76.47} & \textbf{76.47} & 17.65 \\
OLH (Medium)      & 40.63 & 40.63 & \textbf{46.88} & 43.75 & 37.50 & 18.75 \\
OLH (Hard)        & 9.68 & 6.45 & \textbf{25.81} & 16.13 & 19.35 & 12.90 \\
\midrule
Wikipedia (Easy)  & \textbf{86.21} & 68.97 & 72.41 & 75.86 & 79.31 & 27.59 \\
Wikipedia (Medium)& \textbf{56.25} & 53.13 & 43.75 & 50.00 & 37.50 & 9.38 \\
Wikipedia (Hard)  & \textbf{12.12} & \textbf{12.12} & \textbf{12.12} & 9.09 & \textbf{12.12} & 3.03 \\
\midrule
CNN (Easy)        & 35.48 & 41.94 & \textbf{48.39} & 45.16 & 41.94 & 32.26 \\
CNN (Medium)      & 9.68 & \textbf{19.35} & 12.90 & \textbf{19.35} & 9.68 & 19.35 \\
CNN (Hard)        & 3.45 & 6.90 & \textbf{13.79} & 10.34 & 10.34 & 10.34 \\
\midrule
Pitchfork (Easy)  & \textbf{67.74} & 32.26 & 19.35 & 59.38 & 53.13 & 6.25 \\
Pitchfork (Medium)& 25.00 & 9.38 & 12.50 & \textbf{41.18} & 32.35 & 5.88 \\
Pitchfork (Hard)  & \textbf{18.75} & 15.63 & 3.13 & 9.68 & 16.13 & 6.45 \\
\midrule
Petapixel (Easy)  & \textbf{59.38} & 56.25 & 53.13 & 54.84 & 54.84 & 16.13 \\
Petapixel (Medium)& 32.35 & \textbf{47.06} & \textbf{47.06} & 37.50 & 25.00 & 15.63 \\
Petapixel (Hard)  & 22.58 & 19.35 & \textbf{25.81} & 9.38 & 12.50 & 12.50 \\
\midrule
Lonelyplanet (Easy)  & \textbf{37.50} & \textbf{37.50} & 31.25 & 31.25 & 31.25 & 3.13 \\
Lonelyplanet (Medium)& 2.86 & \textbf{11.43} & 2.86 & \textbf{11.43} & 5.71 & 5.71 \\
Lonelyplanet (Hard)  & 3.13 & \textbf{18.75} & 3.13 & 9.38 & 6.25 & 0.00 \\
\bottomrule
\end{tabular}
\caption{Accuracies (\%) across querysets and difficulty levels. Bold values indicate the highest accuracy per dataset–difficulty pair. Model abbreviations: DR = DeepResearcher; CKP = CognitiveKernel-Pro. Queryset abbreviation: OLH = Open Library of Humanities.}
\label{tab:accuracy-transposed}
\end{table*}

\subsection{Experiment settings}


We conducted a systematic evaluation of three mainstream closed-source LLMs with web search capabilities, as well as three recently proposed high-performance open-source agent frameworks. To establish a human-performance benchmark, we uniformly sampled 84 query instances from our benchmark dataset, ensuring balanced representation across different target websites and difficulty levels, and invited domain experts to provide manual answers. Each expert was given a time limit of 15 minutes per query; if the task was not completed within this time, the query was marked as a failure.

The closed-source models evaluated are GPT-4o \citep{hurst_gpt-4o_2024}, Gemini 2.5-flash \citep{google_gemini_2025}, and Perplexity Sonar \citep{perplexityai_models_2025}, each equipped with its official search tool. For open-source agents, we tested Search-R1 \citep{jin_search-r1_2025}, CognitiveKernel-Pro \citep{fang_cognitive_2025}, and DeepResearcher \citep{zheng_deepresearcher_2025}, using the backbone models trained and introduced in their original papers.

To ensure fairness in the experiments, we performed minimal adaptations to certain systems when necessary. For instance, the default search module of Search-R1 does not return web page URLs; therefore, we slightly modified its output format to explicitly include URL information. This adjustment ensures that all agents participating in the evaluation have consistent capabilities and access to the same information during web retrieval and content verification stages.

\begin{table*}
\centering
\small
\begin{tabular}{lcc}
\toprule
\textbf{Difficulty} & \textbf{Completion rate (\%)} & \textbf{Avg. completion time} \\
\midrule
Easy   & 85.7 & 5 min 27 s \\
Medium & 78.6 & 5 min 29 s \\
Hard   & 78.6 & 6 min 41 s \\
\bottomrule
\end{tabular}
\caption{Completion rate and average time consumed by human experts. Only correct answers obtained within 15 minutes are counted.}
\label{tab:human_performance}
\end{table*}

\begin{figure*}[p]
\begin{center}
\includegraphics[width=\linewidth]{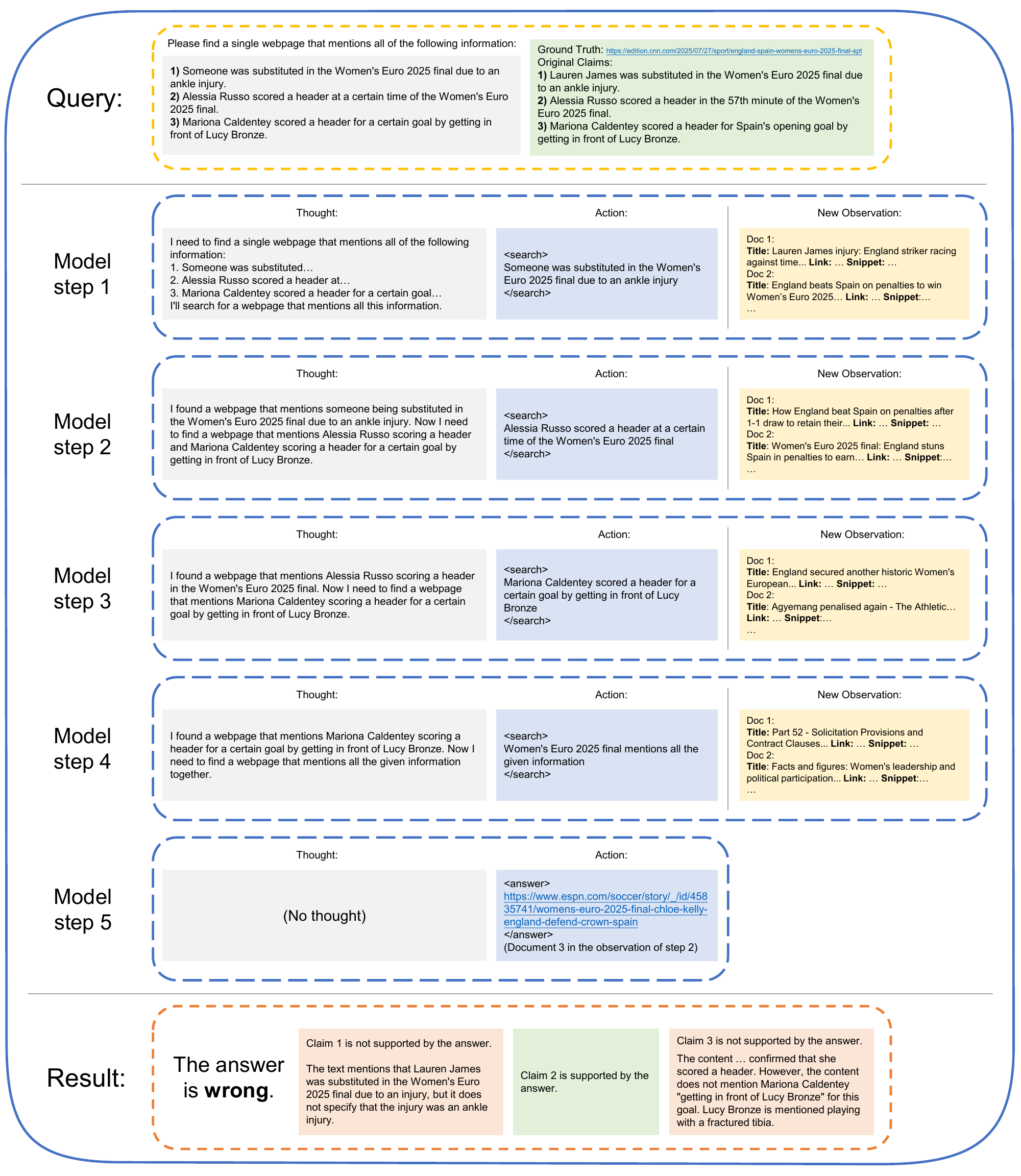}
\end{center}
\caption{A real example illustrating the typical errors that agents exhibit. Due to space limits, some contents were abbreviated using ellipses.}
\label{fig_model_error}
\end{figure*}

\subsection{General discussion}

Table~\ref{tab:accuracy-overview} and~\ref{tab:accuracy-transposed} show the accuracy data of the experiment. We demonstrate our main findings as follows:


\textbf{The semantic relevance between the query and the content significantly affects search performance.} Across all agents, accuracy consistently declines as query difficulty increases. This observation corroborates our experimental design discussed in Sec.~\ref{sec:design}, namely, that agents successful in the Complex Reasoning Search paradigm still struggle with the fuzzy exploratory search paradigm.

\textbf{Our query design effectively excluded the possibility of multiple valid answers.} Although it is not guaranteed that the queries have only one correct answer, we claim that they seldom have multiple answers. In the experiment where 6 models respectively attempted to solve 663 queries, only 7 attempts managed to find a valid answer distinct from the expected correct one, which have also been recognized as correct answer by our validation process.

\textbf{Agents perform very differently on queries with different source websites.} Agent performance varies substantially across source websites. Most agents perform better on academic sources such as ArXiv, Open Library of Humanities, and Wikipedia, and worse on daily-life websites. This discrepancy likely arises because academic sites are more structurally consistent and less cluttered with irrelevant content.


\textbf{Closed-source models are more query-efficient compared to open-source ones.} Despite similar accuracy between open-source and closed-source models, closed-source models were notably more query-efficient. They typically required only one or two search calls per query, whereas open-source agents often needed more than five. This reflects differences in search tool integration that closed systems employ well-tuned search engines, while most open-source agents rely on generic APIs with minimal optimization for relevance.

\subsection{Agent Behavioral Analysis}

During the experiment, we collected the intermediate steps available and analyzed them separately for each agent.

\textbf{Obtaining web contents remains a challenge for open-source agents.} Search-R1 uses basic python libraries, i.e. aiohttp and BeautifulSoup, to acquire HTML of the webpage and parse it into more readable texts. CognitiveKernel-Pro uses Playwright, an automated browser toolkit to get web environment observations. DeepResearcher also uses two Python libraries, i.e. requests and markdownify to get HTML content and convert it into markdown format. In the experiment, all three open-source agents frequently failed at obtaining the full contents of a webpage, returning empty responses or error messages. Therefore, using simplistic HTML extraction methods does not suffice for the real web environment. This problem severely hinders the application potential of such agents, since it also occurs frequently on widely visited websites including CNN and Arxiv.

\textbf{Chunking documents hinders agent's performance on our tasks.} Search-R1 employs a simple chunking strategy. For each webpage in the search results, it separates the full content into several pieces and only returns the pieces containing part or all of the result snippet. While this approach leverages the powerful semantic matching ability of search engines, not providing the whole page to the agent prevents it from successfully solving the query, resulting in cases where the correct result is shown in search results list but not returned in the final answer. Similarly, these cases could also be seen in the intermediate steps of Sonar. This highlights that despite having significant advantages, providing only chunked document is not suitable for all types of tasks.

\textbf{Agents misunderstand the capabilities of search tools.} In the experiment, the open-source agents often misunderstand the function of search tools. Firstly, it confuses global search with domain-restricted search. For instance, after the agent performed a broad search and found multiple webpages satisfying part of the query's criteria, it may wish to narrow down the results, filtering out results satisfying more of the criteria. While this reasoning aligns with humans, it mistakenly performed another global search using another set of keywords, obtaining largely unrelated results. Secondly, it interacts with the search engine in a wrong way. An example is illustrated in Fig.~\ref{fig_model_error}, after the agent obtained results satisfying different parts of the whole criteria separately, it wishes to find a webpage mentioning all required information. Then, it directly searches for \textit{all the given information}, which is of no help at all.

\textbf{Agents may misunderstand the notion of semantic matching.} While the queries ask for webpages that mention certain information, they never require that the answer webpage directly include the description string in its contents. However, the agent sometimes understands the query to bear the latter meaning, excluding correct answers due to them not having the exact description string in their contents. This problem is particularly evident in CognitiveKernel-Pro's intermediate steps. It is the only agent in the three open-source agents that can execute python code, and it frequently attempts to use string matching to determine whether a webpage contains the required information. Because the queries never mention any part of the original text in the ground truth webpage, these attempts always fail. This also accounts for the particularly low accuracy it achieves.

\section{Discussion}

\benchname~introduces a set of queries that share the form of common factoid questions yet have distinct requirements. Even agents that excel on prior benchmarks struggled when confronted with them. This contrast underlines a critical gap in existing search-dedicated benchmarks, namely that they overlook ambiguous user queries and focus on retrieving short factoid answers. In our benchmark, however, agents must retrieve an entire webpage matching vague criteria, which is more reflective of real-world exploratory information seeking.

\textbf{The deliberately ambiguous queries often mislead agents with fragmented search results, and a significant maturity gap exists between proprietary and open-source models in search tool efficiency.} Each piece of information mentioned in the queries is deliberately modified so that there exist on the web many webpages that correspond to it. Agents frequently commit to fractured snippets returned by the search tool, without carefully checking each candidate to identify the correct answer. Differences in search tool utilization further reveal maturity gaps between proprietary and open frameworks. Closed-source models demonstrate greater query efficiency, while open agents engage in inefficient, repetitive searches due to poor planning and misunderstanding of search APIs.

\textbf{Successfully answering~\benchname~queries requires both the ability to judge whether sufficient information has been gathered and to interact with tools appropriately—capabilities lacking in current open-source agents, while even high-performing closed-source models can fail under insufficient context.}  Contemporary search agents lack the ability to explore more information when the context is insufficient~\citep{joren_sufficient_2024}. Another crucial capability is to interact with tools in a reasonable manner. Current search agents misuse search tools, using as keyword abstract commands instead of concrete description of desired information.

\section{Conclusion}

We present \benchname, a benchmark specifically crafted to evaluate how well modern LLM-based search systems behave in retrieving targeted webpages in response to fuzzy exploratory queries. While prior evaluation frameworks stick to a Complex Reasoning Search paradigm, \benchname~represents Fuzzy Exploratory Search tasks. The agent must contend with vague, open-ended queries and deliver an entire webpage that matches implicit criteria, rather than a single factual answer. By deriving from real-world web content 663 queries across seven diverse domains and three difficulty levels, we ensure that the evaluation captures a wide spectrum of challenges.
Empirical results showcase crucial limitations in current systems. State-of-the-art closed-source models, as well as advanced open-source frameworks, only achieve an overall accuracy lower than 35\%, and exhibit inconsistent performance across domains. The findings highlight systemic weaknesses in handling ambiguity, understanding semantic matching, and tool use.
\benchname~exposes a critical gap between current retrieval-augmented systems and the demands of real-world exploratory search. We envision the benchmark guiding development toward agents that are uncertainty-aware, semantically robust, and capable of verifying retrieved content against vague criteria. Its modular design further allows expansion to new domains, languages, and modalities, ensuring lasting relevance. Ultimately, advancing performance on this benchmark will be central to building web agents that can reason and search with the flexibility and persistence characteristic of human exploratory inquiry.

\section{Limitations}

In spite of the diverse domains \benchname~spans, it only represents a limited slice of the real web environment. Certain content types including highly interactive sites, social media, or non-English resources remain underrepresented. Furthermore, the benchmark relies on a fixed corpus of queries and webpages, capturing the web's state at a particular time. As the web evolves, results may become outdated, and performance improvements might reflect data familiarity rather than genuine reasoning progress. Lastly, some performance disparities stem from toolchain limitations (e.g., web scraping errors, incomplete page rendering) rather than reasoning ability. This technical dependency complicates fair comparison across frameworks.

\bibliography{needle_in_the_web}

\appendix

\section{Websites Chosen for Query Generation}

The queries in \benchname~were generated based on contents collected from seven distinct websites. We briefly explain here why we adopted this setting, and how each of the websites was chosen.

Our query collection pipeline needs well-formed contents. Considering that dealing with raw HTML could be tedious and error-prone, we used FireCrawl to obtain webpage contents in Markdown format. Furthermore, we individually wrote scripts for each website to exclude irrelevant texts (e.g. advertisements) from the contents. Because the scripts are specifically tailored according to the website, we limited the source webpages within seven domains. 

The length of the contents is also important. If the content is too short, there would be insufficient information for generating the query; If too long, our setting of presenting three factual statements in the query may not be able to reliably locate the source webpage. Therefore, we chose all websites to feature articles of length falling into an appropriate range.

ArXiv (https://arxiv.org/) is a free, open-access repository for research papers, primarily in physics, mathematics, computer science, quantitative biology, quantitative finance, and statistics. The Open Library of Humanities (https://www.openlibhums.org/) is a nonprofit, open-access publishing platform that supports and publishes scholarly journals in the humanities. We chose these two websites to represent the academic and professional side of the web. Specifically, we sampled webpages uniformly across the latest papers in each subcategory of ArXiv Computer Science Repository and OLH. Because the whole research article is too long for our query collection pipeline, we only used the abstract and introduction sections of each article.

Wikipedia (https://simple.wikipedia.org/) is a free, online encyclopedia that anyone can edit, containing millions of articles in multiple languages, written collaboratively by volunteers around the world. We chose Wikipedia as one of the queries' domains since it covers a wide range of topics, and its contents are relatively well-formed. We leveraged an API it provided for randomly sampling pages written in English.

The other four websites represent the kind of websites browsed in daily life for information or entertainment. For these websites, we simply collected the latest articles from them. CNN (Cable News Network, https://edition.cnn.com/) is a major American news organization founded in 1980 by Ted Turner. It is known for breaking news coverage, political reporting, and live global event broadcasting. Pitchfork (https://pitchfork.com/) is an online music publication known for its in-depth reviews, artist interviews, and music news, covering a wide range of genres with a focus on indie, alternative, and experimental music. PetaPixel (https://petapixel.com/) is a popular online publication that focuses on photography news, camera gear reviews, and industry commentary. Lonely Planet (https://www.lonelyplanet.com/articles) is a renowned travel guidebook publisher and website that provides information and recommendations for travelers worldwide.

\section{Agent Trajectory Examples}

\subsection{An example query of medium difficulty from the Arxiv queryset}

\begin{lstlisting}
Please find a single webpage that mentions all of the following information:

1. On the CMMD dataset, GPT-5 reached a certain percentage abnormality detection,
2. Performance in mammogram interpretation can vary significantly even among certain people,
3. Certain models further extend advances in AI for medical imaging.
\end{lstlisting}

\subsection{Search-R1's agent trajectory}

\begin{lstlisting}
"<think>I need to find a single webpage that mentions all of the following information:
1. On the CMMD dataset, GPT-5 reached a certain percentage abnormality detection.
2. Performance in mammogram interpretation can vary significantly even among certain people.
3. Certain models further extend advances in AI for medical imaging.

I'll search for it.</think>

<search> On the CMMD dataset, GPT-5 reached a certain percentage abnormality detection. Performance in mammogram interpretation can vary significantly even among certain people. Certain models further extend advances in AI for medical imaging </search>
",
"Doc 1(Title: \"The Role of AI in Hospitals and Clinics: Transforming Healthcare in ...\", Link: https://www.mdpi.com/2306-5354/11/4/337) As healthcare systems around the world face challenges such as escalating costs, limited access, and growing demand for personalized care, artificial intelligence (AI) is emerging as a key force for transformation. This review is motivated by the urgent need to harness AI\u2019s potential to mitigate these issues and aims to critically assess AI\u2019s integration in different healthcare domains. We explore how AI empowers clinical decision-making, optimizes hospital operation and management, refines medical image analysis, and revolutionizes patient care and monitoring through AI-powered wearables. Through several case studies, we review how AI has transformed specific healthcare domains and discuss the remaining challenges and possible solutions. Additionally, we will discuss methodologies for assessing AI healthcare solutions, ethical challenges of AI deployment, and the importance of data privacy and bias mitigation for responsible technology use. By presenting a critical assessment of AI\u2019s transformative potential, this review equips researchers with a deeper understanding of AI\u2019s current and future impact on healthcare. It encourages an interdisciplinary dialogue between researchers, clinicians, and technologists to navigate the complexities of AI implementation, fostering the development of AI-driven solutions that prioritize ethical standards, equity, and a patient-centered approach.
Doc 2(Title: \"Deep Learning in Digital Breast Tomosynthesis: Current Status ...\", Link: https://pmc.ncbi.nlm.nih.gov/articles/PMC12146671/) Breast disease classification and diagnosis, lesion segmentation and detection, and medical image generation can be achieved on DBT data via DL models. The integration of image data obtained from DBT is essential for developing an intelligent early breast cancer screening system based on DL. A comprehensive and in\u2010depth exploration of the application of DL in the field of DBT is particularly necessary to fully realize its potential.
Doc 3(Title: \"Deep Learning in Breast Cancer Imaging: State of the Art and ...\", Link: https://www.mdpi.com/2075-4418/14/8/848) No snippet available.
Doc 4(Title: \"JMIR Preprints: JP\", Link: https://preprints.jmir.org/) No snippet available.
Doc 5(Title: \"Xiaofeng YANG | Professor | Ph.D | Emory University, Atlanta | EU ...\", Link: https://www.researchgate.net/profile/Xiaofeng-Yang-19) No snippet available.
",
"<think>I found a relevant webpage that mentions some of the information. Now I need to find a webpage that mentions all the information.</think>

<search> On the CMMD dataset, GPT-5 reached a certain percentage abnormality detection </search>

",
"Doc 1(Title: \"GPT-5 Demonstrates Mammography VQA Performance On BI-RADS ...\", Link: https://quantumzeitgeist.com/gpt-5-demonstrates-mammography-vqa-performance-on-bi-rads-assessment-and-malignancy-classification/) The research investigates how large language models, specifically GPT-5, can be applied to medical image analysis, focusing on mammography datasets including EMBED, InBreast, CMMD, and CBIS-DDSM. These tasks, BI-RADS assessment, abnormality detection, and malignancy classification, are critical components of breast cancer diagnosis and screening. Evaluations focused on the model\u00e2\u20ac\u2122s ability to classify breast density, detect abnormalities, and determine if a finding is likely to be malignant. Across the EMBED dataset, GPT-5 achieves 56. 8% accuracy in density assessment, 52.5% in distortion analysis, 64. 5% in mass classification, 63. 5% in calcification classification, and 52. 8% in malignancy classification. Across the datasets, GPT-5 achieved BI-RADS accuracy ranging from 36.9% to 69. 3%, abnormality detection rates from 32. 3% to 66. 0%, and malignancy classification accuracy from 35. 0% to 58.2%. While these results represent a significant improvement over previous models, GPT-5 still lags behind the performance of human experts in both sensitivity and specificity, achieving 63. 5% and 52. 3% respectively.
Doc 2(Title: \"Examples include (top) a 49-year-old White woman with right ...\", Link: https://www.researchgate.net/figure/Examples-include-top-a-49-year-old-White-woman-with-right-mediolateral-oblique_fig3_366870100) No snippet available.
Doc 3(Title: \"AI-Driven Automated Blood Cell Anomaly Detection: Enhancing ...\", Link: https://www.mdpi.com/2313-433X/11/5/157) No snippet available.
Doc 4(Title: \"Variations in shape and appearance of polyps in colonoscopy videos.\", Link: https://www.researchgate.net/figure/ariations-in-shape-and-appearance-of-polyps-in-colonoscopy-videos_fig1_297659797) No snippet available.
Doc 5(Title: \"Editor's Choice -- European Society for Vascular Surgery (ESVS ...\", Link: https://www.ejves.com/article/S1078-5884(23)00889-4/fulltext) No snippet available.
",
"<think>I found a webpage that mentions On the CMMD dataset, GPT-5 reached a certain percentage abnormality detection. Now I need to find a webpage that mentions the other two pieces of information.</think>

<search> Performance in mammogram interpretation can vary significantly even among certain people </search>

",
"Doc 1(Title: \"Artificial Intelligence in Breast Cancer Diagnosis and Personalized ...\", Link: https://pmc.ncbi.nlm.nih.gov/articles/PMC10625863/) AI triaging is another method for testing AI algorithms. Because the majority of screening mammograms are negative for malignancy, removing even a portion of normal examinations can significantly reduce the workload. Dembrower et al. [29] showed that AI can be set at a threshold at which 60% of the cases can be safely removed from the worklist without risking missing cancer cases. Other studies have reported similar results, with a 47% reduction in workload, resulting in only 7% missed cancers [58]. Additionally, a \u201crule-in\u201d approach can be employed, where cases labeled as benign by human readers but assigned a high score by AI are automatically recalled for further testing. This workflow, combined with the \u201crule-out\u201d approach, can significantly reduce the workload, while increasing the detection of subsequent interval cancers (ICs) and next-round detected cancers [29].
Doc 2(Title: \"Improving radiologist's ability in identifying particular abnormal ...\", Link: https://www.nature.com/articles/s41598-021-89214-3) No snippet available.
Doc 3(Title: \"The Role of AI in Hospitals and Clinics: Transforming Healthcare in ...\", Link: https://pmc.ncbi.nlm.nih.gov/articles/PMC11047988/) AI in mental health: In the field of mental health, AI is used to personalize treatment approaches. By monitoring patterns in speech [51], behavior [52], and social media activity [53,54], AI tools can help in identifying the onset of mental health issues and suggest interventions tailored to the individual\u2019s unique situation. This personalized approach is crucial in mental health, where treatment efficacy can vary significantly from person to person.
In future research and development within mental health treatment, a promising direction is the integration of AI systems with emotional intelligence [55]. Such systems could be crucial in early detection and intervention of mental health disorders by analyzing speech and behavior patterns for signs of conditions like depression or anxiety. Further exploration into personalizing therapy using AI could lead to more individualized and effective care.
Doc 4(Title: \"Workshop Summary - Assessing and Improving the Interpretation of ...\", Link: https://www.ncbi.nlm.nih.gov/books/NBK327233/) Millions of women undergo screening mammography regularly with the hope of detecting breast cancer at an earlier and more curable stage. But the ability of such screening to accurately detect early cancers depends on the quality of mammography, including high-quality image acquisition and interpretation. To help ensure the quality of mammography, Congress passed the Mammography Quality Standards Act (MQSA) in 1994 and last reauthorized it in 2004. In advance of its expected reauthorization in 2007, Congress requested a consensus study from the Institute of Medicine (IOM) recommending ways to improve the quality of mammography, with an emphasis on image interpretation. The resulting report, Improving Breast Imaging Quality Standards, highlighted the need to decrease variability in mammography interpretation in the United States and identified gaps in the evidence needed to develop best practices (IOM, 2005). The consensus committee found that while the technical quality of mammography had improved since MQSA implementation, mammography interpretation remained quite variable, and that this variability limited the full potential of mammography to reduce breast cancer mortality by detecting breast cancers at an early stage.
Doc 5(Title: \"Using global feedback to induce learning of gist of abnormality in ...\", Link: https://cognitiveresearchjournal.springeropen.com/articles/10.1186/s41235-022-00457-8) By comparing human and machine performance on specific images, we can learn more about whether these models capture the same image features that humans might be using\u2014which in turn can be informative for human perception. The single breast classifier (SBC) version of Wu et al. (2019) deep neural network (DNN) for breast cancer screening predicts the probability of both benign and malignant abnormalities for individual unilateral mammograms and reaches a high performance (AUC malignant: 0.84\u20130.90, AUC benign: 0.74\u20130.76) on detecting visible abnormalities in a large screening data set, which make it suitable for our purposes. We will use both the SBC and SBC heatmap (SBC\u2009+\u2009HM) version, which adds heatmaps generated via a secondary network which examines smaller pixel patches for their malignancy probability. These heatmaps provide additional scrutiny of local information that is expected to improve performance, while the SBC without heatmaps would be more equivalent to the global information used in gist extraction. Comparing the probability scores from both the SBC and SBC\u2009+\u2009HM network to human rating scores will allow us to investigate whether they capture similar information used by human gist extraction of medical abnormality.
Furthermore, the observed variability in learning rates between participants in this study matches the previous literature. Learning rates differ significantly between individuals across seven perceptual tasks in the visual and auditory domain, such as Vernier acuity, face view discrimination, and auditory frequency discrimination (Yang et al., 2020). Importantly, the contribution of participant-specific (36.8%) factors is approximately equal to the task-specific (~\u200938.6%) factors influencing learning rate, underlining the large impact individual differences can have on learning rates across tasks. Individual differences in learning rates have also been demonstrated in spatial learning in virtual environments (Waller, 2000).
",
"<think>I found a webpage that mentions Performance in mammogram interpretation can vary significantly even among certain people. Now I need to find a webpage that mentions Certain models further extend advances in AI for medical imaging.</think>

<search> Certain models further extend advances in AI for medical imaging </search>

",
"Doc 1(Title: \"Microsoft collaborates with Mass General Brigham and University of ...\", Link: https://news.microsoft.com/source/2024/07/24/microsoft-collaborates-with-mass-general-brigham-and-university-of-wisconsin-madison-to-further-advance-ai-foundation-models-for-medical-imaging/) No snippet available.
Doc 2(Title: \"The Role of AI in Hospitals and Clinics: Transforming Healthcare in ...\", Link: https://pmc.ncbi.nlm.nih.gov/articles/PMC11047988/) No snippet available.
Doc 3(Title: \"A systematic review of deep learning data augmentation in medical ...\", Link: https://www.sciencedirect.com/science/article/pii/S277244252400042X) No snippet available.
Doc 4(Title: \"How Artificial Intelligence Is Shaping Medical Imaging Technology ...\", Link: https://pmc.ncbi.nlm.nih.gov/articles/PMC10740686/) The attention mechanism is a fundamental component in transformers. It plays a pivotal role in enabling the model to selectively focus on different parts of the input data with varying degrees of attention. At its core, the attention mechanism allows the model to assign varying levels of importance to different elements within the input data. This means the model can \u201cpay attention\u201d to specific elements while processing the data, prioritizing those that are most relevant to the task at hand. This selective attention enhances the model\u2019s ability to capture essential information and relationships within the input. The mechanism operates as follows: First, the input data is organized into a sequence of elements, such as tokens in a sentence for NLP or patches in an image for computer vision. Then, the mechanism introduces three sets of learnable parameters: query (Q), key (K), and value (V). The query represents the element of interest, while the key and value pairs are associated with each element in the input sequence. For each element in the input sequence, the attention mechanism calculates an attention score, reflecting the similarity between the query and the key for that element. The method used to measure this similarity can vary, with techniques like dot product and scaled dot product being common choices. These attention scores represent how relevant each element is to the query. The next step involves applying the softmax function to the attention scores. This converts them into weights that sum to one, effectively determining the importance of each input element concerning the query. The higher the weight, the more attention the model allocates to that specific element in the input data. Finally, the attention mechanism computes a weighted sum of the values, using the attention weights. The resulting output is a combination of information from all input elements, with elements more relevant to the query receiving higher weight in the final representation [36,37].
Doc 5(Title: \"Revolutionizing healthcare: the role of artificial intelligence in clinical ...\", Link: https://bmcmededuc.biomedcentral.com/articles/10.1186/s12909-023-04698-z) No snippet available.
",
"<think>I found a webpage that mentions Certain models further extend advances in AI for medical imaging. Now I need to find a webpage that mentions all the information.</think>

<search> No source found. </search>

",
"Doc 1(Title: \"Projector keeps saying \u201cno source found\u201d : r/hometheater\", Link: https://www.reddit.com/r/hometheater/comments/1ev24ew/projector_keeps_saying_no_source_found/) No snippet available.
Doc 2(Title: \"command line - Error message 'source: not found' when running a ...\", Link: https://askubuntu.com/questions/504546/error-message-source-not-found-when-running-a-script) No snippet available.
Doc 3(Title: \"Real debrid + fen no sources found : r/Addons4Kodi\", Link: https://www.reddit.com/r/Addons4Kodi/comments/1atqae9/real_debrid_fen_no_sources_found/) No snippet available.
Doc 4(Title: \"No source list found, or it is empty - Help - Pi-hole Userspace\", Link: https://discourse.pi-hole.net/t/no-source-list-found-or-it-is-empty/62554) Restore the default adlist by adding the following URL to your adlists, saving the change, and then rebuilding gravity.
https://raw.githubusercontent.com/StevenBlack/hosts/master/hosts
Doc 5(Title: \"PS5 no source found on Optoma UHD35X : r/projectors\", Link: https://www.reddit.com/r/projectors/comments/11beihn/ps5_no_source_found_on_optoma_uhd35x/) No snippet available.
",
\end{lstlisting}

\subsection{DeepResearcher's agent trajectory}

\begin{lstlisting}
<|im_start|>system
## Background information 
* Today is 2025-09-23
* You are Deep AI Research Assistant

The question I give you is a complex question that requires a *deep research* to answer.

I will provide you with two tools to help you answer the question:
* A web search tool to help you perform google search. 
* A webpage browsing tool to help you get new page content.

You don't have to answer the question now, but you should first think about the research plan or what to search next.

Your output format should be one of the following two formats:

<think>
YOUR THINKING PROCESS
</think>
<answer>
YOUR ANSWER AFTER GETTING ENOUGH INFORMATION
</answer>

or

<think>
YOUR THINKING PROCESS
</think>
<tool_call>
YOUR TOOL CALL WITH CORRECT FORMAT
</tool_call>

You should always follow the above two formats strictly.
Only output the final answer (in words, numbers or phrase) inside the <answer></answer> tag, without any explanations or extra information. If this is a yes-or-no question, you should only answer yes or no.


# Tools

You may call one or more functions to assist with the user query.

You are provided with function signatures within <tools></tools> XML tags:
<tools>
{\"type\": \"function\", \"function\": {\"name\": \"web_search\", \"description\": \"Search the web for relevant information from google. You should use this tool if the historical page content is not enough to answer the question. Or last search result is not relevant to the question.\", \"parameters\": {\"type\": \"object\", \"properties\": {\"query\": {\"type\": \"array\", \"items\": {\"type\": \"string\", \"description\": \"The query to search, which helps answer the question\"}, \"description\": \"The queries to search\"}}, \"required\": [\"query\"], \"minItems\": 1, \"uniqueItems\": true}}}
{\"type\": \"function\", \"function\": {\"name\": \"browse_webpage\", \"description\": \"Browse the webpage and return the content that not appeared in the conversation history. You should use this tool if the last action is search and the search result maybe relevant to the question.\", \"parameters\": {\"type\": \"object\", \"properties\": {\"url_list\": {\"type\": \"array\", \"items\": {\"type\": \"string\", \"description\": \"The chosen url from the search result, do not use url that not appeared in the search result\"}, \"description\": \"The chosen urls from the search result.\"}}, \"required\": [\"url_list\"]}}}
</tools>

For each function call, return a json object with function name and arguments within <tool_call></tool_call> XML tags:
<tool_call>
{\"name\": <function-name>, \"arguments\": <args-json-object>}
</tool_call><|im_end|>
<|im_start|>user
Please find a single webpage that mentions all of the following information:

On the CMMD dataset, GPT-5 reached a certain percentage abnormality detection.
Performance in mammogram interpretation can vary significantly even among certain people.
Certain models further extend advances in AI for medical imaging.

Your response will be parsed by a program, so make sure to observe the formatting instructions! You need to format your response as follows:
<answer>the url of the webpage that you found</answer>
...
Make sure to explicitly include `<answer>` and `</answer>` with surrounding angle brackets in your response, even if you do not have an answer.
If you are unable to find the webpage that mentions all the information, return the following:
<answer> No source found. </answer>
<|im_end|>
<|im_start|>assistant
<think> 
To find a webpage that mentions all the requested information, I should first perform a web search using the given keywords.
</think>
<tool_call>
{\"name\": \"web_search\", \"arguments\": {\"query\": [\"CMMD dataset GPT-5 abnormality detection\", \"mammogram interpretation performance\", \"AI for medical imaging advances\"]}}
</tool_call><|im_end|>
<|im_start|>user
<tool_response>
[{'search_query': 'CMMD dataset GPT-5 abnormality detection', 'web_page_info_list': [{'quick_summary': 'On CMMD, GPT-5 reached 32.3% abnormality detection and 55.0% malignancy accuracy. On CBIS-DDSM, it achieved 69.3% BI-RADS accuracy, 66.0% ...', 'title': '[2508.11628] Is ChatGPT-5 Ready for Mammogram VQA? - arXiv', 'url': 'https://arxiv.org/abs/2508.11628'}, {'quick_summary': 'In CMMD, GPT-5 records 32.3% abnormality detection and 55.0% malignancy accuracy, versus 79.7% from HybMNet. Report issue for preceding element.', 'title': 'Is ChatGPT-5 Ready for Mammogram VQA? - arXiv', 'url': 'https://arxiv.org/html/2508.11628v1'}, {'quick_summary': 'Across the datasets, GPT-5 achieved BI-RADS accuracy ranging from 36.9% to 69. 3%, abnormality detection rates from 32. 3% to 66. 0%, and ...', 'title': 'GPT-5 Demonstrates Mammography VQA Performance On BI-RADS ...', 'url': 'https://quantumzeitgeist.com/gpt-5-demonstrates-mammography-vqa-performance-on-bi-rads-assessment-and-malignancy-classification/'}, {'quick_summary': 'On CMMD, GPT-5 reached 32.3% abnormality detection and 55.0% malignancy accuracy.', 'title': '(PDF) Is ChatGPT-5 Ready for Mammogram VQA? - ResearchGate', 'url': 'https://www.researchgate.net/publication/394525007_Is_ChatGPT-5_Ready_for_Mammogram_VQA'}, {'quick_summary': 'On CMMD, GPT-5 reached 32.3% abnormality detection and 55.0% malignancy accuracy. On CBIS-DDSM, it achieved 69.3% BI-RADS accuracy, 66.0% ...', 'title': '(PDF) Capabilities of GPT-5 on Multimodal Medical Reasoning', 'url': 'https://www.researchgate.net/publication/394439108_Capabilities_of_GPT-5_on_Multimodal_Medical_Reasoning'}, {'quick_summary': 'Mammography, a noninvasive imaging tool with low cost, is widely used to diagnose breast disease at an early stage due to its high sensitivity.', 'title': 'The Chinese Mammography Database (CMMD)', 'url': 'https://wiki.cancerimagingarchive.net/pages/viewpage.action?pageId=70230508'}, {'quick_summary': 'The deep learning models used for breast lesion segmentation and detection on DBT data include U\u2010Net and GCN. ... anomaly detection [205]. A ...', 'title': 'Deep Learning in Digital Breast Tomosynthesis - PubMed Central', 'url': 'https://pmc.ncbi.nlm.nih.gov/articles/PMC12146671/'}, {'quick_summary': '... datasets (EMBED, InBreast, CMMD, CBIS-DDSM) for tasks like BI-RADS assessment, abnormality detection, and malignancy classification. While *GPT- ...', 'title': 'Is ChatGPT-5 Ready for Mammogram VQA - YouTube', 'url': 'https://www.youtube.com/watch?v=EZXdwtqbmJw'}, {'quick_summary': 'Summary. The TOMPEI-CMMD dataset adds the following image analyses to the original CMMD dataset on TCIA:.', 'title': 'TOMPEI-CMMD - The Cancer Imaging Archive (TCIA)', 'url': 'https://www.cancerimagingarchive.net/analysis-result/tompei-cmmd/'}, {'quick_summary': 'Among them, 798 breasts from 728 patients with abnormal AI results (abnormality score \u2265 10 generated by an AI-CAD explained below) in screening and diagnostic ...', 'title': 'Positive Predictive Values of Abnormality Scores From a ...', 'url': 'https://pmc.ncbi.nlm.nih.gov/articles/PMC10973732/'}]}, {'search_query': 'mammogram interpretation performance', 'web_page_info_list': [{'quick_summary': 'These metrics were designed to evaluate radiologist performance in breast imaging interpretation, yet they are also widely used to inform women, healthcare ...', 'title': 'New mammography screening performance metrics based on the ...', 'url': 'https://pmc.ncbi.nlm.nih.gov/articles/PMC7319901/'}, {'quick_summary': 'This narrative review aims to identify what factors are linked to diagnostic performance variation for those who interpret mammograms.', 'title': 'Identification of factors associated with diagnostic performance ...', 'url': 'https://www.sciencedirect.com/science/article/pii/S1078817423000044'}, {'quick_summary': 'This study identif ed minimally acceptable performance levels for interpreters of screening mammography studies. Interpret- ing physicians whose performance ...', 'title': '[PDF] Identifying Minimally Acceptable Interpretive Performance Criteria ...', 'url': 'https://www2.rsna.org/timssnet/radiologyselect/breastcancer/PDF%20files/Category%201/Carney.pdf'}, {'quick_summary': \"The availability of previous screening mammograms improves radiographers' ability to discriminate between normal and abnormal mammograms and reduce the false ...\", 'title': 'Does access to prior mammograms improve the performance of ...', 'url': 'https://www.sciencedirect.com/science/article/pii/S1078817424003560'}, {'quick_summary': 'The purposes of this study were to determine whether US radiologists accurately estimate their own interpretive performance of screening mammography.', 'title': \"Mammographic Interpretation: Radiologists' Ability to Accurately ...\", 'url': 'https://ajronline.org/doi/10.2214/AJR.11.7402'}, {'quick_summary': 'Those performing diagnostic mammography were more likely to achieve acceptable PPV1, PPV2, PPV3, invasive CDR, and CDR (OR, 1.9\u20132.9). Those ...', 'title': 'Radiologist Characteristics Associated with Interpretive Performance ...', 'url': 'https://pubs.rsna.org/doi/abs/10.1148/radiol.2021204379'}, {'quick_summary': 'Radiologists who reported enjoying interpreting screening mammograms were more likely to be women, spend at least 20% of their time in breast imaging, have a ...', 'title': \"Radiologists' Performance and Their Enjoyment of Interpreting ...\", 'url': 'https://ajronline.org/doi/10.2214/AJR.08.1647?doi=10.2214/AJR.08.1647'}, {'quick_summary': \"Mode of Interpretation. Another important factor that can influence the performance characteristics of mammography is a facility's mode of film interpretation.\", 'title': '2 Improving Interpretive Performance in Mammography', 'url': 'https://nap.nationalacademies.org/read/11308/chapter/4'}, {'quick_summary': 'The use of computer-aided detection is associated with reduced accuracy of interpretation of screening mammograms.', 'title': 'Influence of Computer-Aided Detection on Performance of ...', 'url': 'https://www.nejm.org/doi/full/10.1056/NEJMoa066099'}]}, {'search_query': 'AI for medical imaging advances', 'web_page_info_list': [{'quick_summary': 'AI-based diagnostic tools not only speed up the interpretation of complex images but also improve early detection of disease, ultimately delivering better ...', 'title': 'How Artificial Intelligence Is Shaping Medical Imaging Technology', 'url': 'https://pmc.ncbi.nlm.nih.gov/articles/PMC10740686/'}, {'quick_summary': 'We develop AI methods to reconstruct images from accelerated MRI scans, with the aim of making MRI 10 times faster. Opens in a new tab. Our algorithms learn a ...', 'title': 'Artificial Intelligence in Biomedical Imaging | NYU Langone Health', 'url': 'https://med.nyu.edu/departments-institutes/radiology/research/ai-biomedical-imaging'}, {'quick_summary': \"AI tools can alleviate radiologists' workload by automating mundane tasks, reducing burnout and enhancing care delivery in underserved areas. At ...\", 'title': \"The Future of Radiology: AI's Transformative Role in Medical Imaging\", 'url': 'https://www.rsna.org/news/2025/january/role-of-ai-in-medical-imaging'}, {'quick_summary': 'AI algorithms can quickly analyze large amounts of imaging data, identifying patterns and abnormalities that may be overlooked by human eyes.', 'title': 'Artificial Intelligence in Medical Imaging - Spectral AI', 'url': 'https://www.spectral-ai.com/blog/artificial-intelligence-in-medical-imaging/'}, {'quick_summary': 'AI can be particularly useful in medical imaging solutions, enhancing processes such as screening, patient eligibility, and disease detection and response ...', 'title': 'The rise and role of AI in medical imaging - ICON plc', 'url': 'https://www.iconplc.com/insights/blog/2024/06/14/rise-and-role-ai-medical-imaging'}, {'quick_summary': 'AI algorithms can analyse medical images with remarkable accuracy and speed, often surpassing human capabilities.', 'title': 'The Good, the Bad, and the Ugly of AI in Medical Imaging - EMJ', 'url': 'https://www.emjreviews.com/radiology/article/the-good-the-bad-and-the-ugly-of-ai-in-medical-imaging-j140125/'}, {'quick_summary': 'staff. At the same time, groundbreaking developments in the field of artificial intelligence continue to transform the face of the imaging ...', 'title': 'How AI is transforming Radiology - YouTube', 'url': 'https://www.youtube.com/watch?v=3DUyzPvsMQ8'}, {'quick_summary': 'The future of AI in medical imaging holds the potential to completely transform how we detect diseases like cancer, interpret medical images, and even predict ...', 'title': 'Future of AI in medical imaging: Challenges and opportunities', 'url': 'https://quibim.com/news/ai-in-medical-imaging/'}, {'quick_summary': 'AI-driven medical image analysis systems are booming in the fields of cardiovascular, oncology, respiratory, orthopedics, and ophthalmology fields.', 'title': 'Artificial intelligence and medical imaging - Nature', 'url': 'https://www.nature.com/collections/bjeiihhgfa'}]}]
</tool_response><|im_end|>
<|im_start|>assistant
<think> 
The search results have provided some pages with relevant information. I will now browse the webpages to find the one that contains all the required information.
</think>
<tool_call>
{\"name\": \"browse_webpage\", \"arguments\": {\"url_list\": [\"https://arxiv.org/abs/2508.11628\", \"https://quantumzeitgeist.com/gpt-5-demonstrates-mammography-vqa-performance-on-bi-rads-assessment-and-malignancy-classification/\"]}}
</tool_call><|im_end|>
<|im_start|>user
<tool_response>
[{'information': [{'page_number': 0, 'page_summary': \"- On the CMMD dataset, GPT-5 achieved 32.3% abnormality detection accuracy.\
- The same paper also reports GPT-5's malignancy accuracy on CMMD as 55.0%.\
- GPT-5 was evaluated alongside GPT-4o on four public mammography datasets: EMBED, InBreast, CMMD, and CBIS-DDSM.\
- GPT-5 consistently outperformed GPT-4o but still lagged behind human experts and domain-specific fine-tuned models.\
- The study includes performance metrics for BI-RADS assessment, abnormality detection, and malignancy classification across datasets.\
- The paper explicitly states that GPT-5\u2019s sensitivity was 63.5% and specificity was 52.3% across tasks, indicating variability in performance.\
- The paper highlights that improvements from GPT-4o to GPT-5 show a promising trend for general LLMs in mammography VQA, implying that such models are extending advances in AI for medical imaging.\
- The paper is authored by researchers from multiple institutions and was submitted on August 15, 2025, suggesting it is a recent and credible source.\
- The abstract mentions that performance in mammogram interpretation varies significantly even among human experts, which directly addresses the sub-question about variability among certain people.\"}], 'url': 'https://arxiv.org/abs/2508.11628'}, {'information': [{'page_number': 0, 'page_summary': '- GPT-5 was evaluated on the CMMD dataset as part of a study on mammography analysis.\
- On the CMMD dataset, GPT-5 achieved an abnormality detection rate of 66.0%.\
- The study was conducted by researchers from Emory University\u2019s Winship Cancer Institute and published on August 18, 2025.\
- GPT-5 was benchmarked against GPT-4o and human experts across multiple datasets including CMMD, EMBED, InBreast, and CBIS-DDSM.\
- GPT-5 showed improved performance over GPT-4o but still fell short of human expert accuracy in mammogram interpretation.\
- The study also evaluated BI-RADS assessment and malignancy classification, with GPT-5 achieving 36.9%\u201369.3% accuracy in BI-RADS and 35.0%\u201358.2% in malignancy classification across datasets.\
- The research highlights that performance in mammogram interpretation varies significantly even among human experts, implying human variability is a known factor in the field.\
- The paper suggests that such AI models represent further advances in AI for medical imaging, particularly in visual question answering (VQA) for radiology.'}, {'page_number': 1, 'page_summary': '- The webpage confirms that GPT-5 was evaluated on the CMMD dataset using a visual question answering (VQA) framework, aligning with the previously reported evaluation methodology.\
- It reiterates that GPT-5 outperformed GPT-4o in mammogram interpretation tasks, including abnormality detection, though it still did not reach human expert levels \u2014 consistent with prior context.\
- The page explicitly links the study to the arXiv paper at https://arxiv.org/abs/2508.11628, which is the source of the findings mentioned in the context so far.\
- The webpage adds that the VQA approach was applied across four datasets: EMBED, InBreast, CMMD, and CBIS-DDSM, reinforcing the multi-dataset validation of GPT-5\u2019s performance.\
- It emphasizes that the questions were generated from structured clinical metadata to ensure direct correspondence with verified labels, enhancing methodological rigor \u2014 a detail not previously highlighted.\
- The page also confirms that GPT-5\u2019s performance improvements represent \u201cfurther advances in AI for medical imaging,\u201d particularly through VQA, which directly supports the third component of the main question.\
- The webpage does not introduce any new numerical values for GPT-5\u2019s abnormality detection rate on CMMD beyond the previously reported 66.0%, nor does it contradict existing data.\
- It does not provide additional information on human performance variability beyond what was already stated in the context, though it implicitly supports it by contrasting AI with expert performance.'}], 'url': 'https://quantumzeitgeist.com/gpt-5-demonstrates-mammography-vqa-performance-on-bi-rads-assessment-and-malignancy-classification/'}]
</tool_response><|im_end|>
<|im_start|>assistant
<think> 
From the search results, I can see that the webpage at https://quantumzeitgeist.com/gpt-5-demonstrates-mammography-vqa-performance-on-bi-rads-assessment-and-malignancy-classification/ provides the required information. It mentions the GPT-5's performance on the CMMD dataset, the variability in mammogram interpretation performance among human experts, and how AI models are advancing in medical imaging through VQA.
</think>
<answer>https://quantumzeitgeist.com/gpt-5-demonstrates-mammography-vqa-performance-on-bi-rads-assessment-and-malignancy-classification/</answer><|im_end|>
\end{lstlisting}

\subsection{CognitiveKernel-Pro's agent trajectory}

\begin{lstlisting}
"steps": [
{
"step_idx": 0,
"plan": {
"thought": "Since there are no previous steps or progress state, I will perform a web search to find a webpage that mentions all the required information about the CMMD dataset, GPT-5, mammogram interpretation, and AI advances in medical imaging.",
"code": "{
\"completed_list\": [],
\"todo_list\": [\"Perform a web search to find a webpage that mentions all the required information.\"],
\"experience\": [],
\"information\": []
}",
"state": {
"completed_list": [],
"todo_list": [
"Perform a web search to find a webpage that mentions all the required information."
],
"experience": [],
"information": []
},
"llm_input": [
{
"role": "system",
"content": "You are a strategic assistant responsible for the high-level planning module of the Cognitive Kernel, an initial autopilot system designed to accomplish user tasks efficiently.

## Available Information
- `Target Task`: The specific task to be completed.
- `Recent Steps`: The most recent actions taken by the agent.
- `Previous Progress State`: A JSON representation of the task's progress, including key information and milestones.
- `Sub-Agent Functions` and `Tool Functions`: Definitions of available sub-agents and tools for task execution.

## Progress State
The progress state is crucial for tracking the task's advancement and includes:
- `completed_list` (List[str]): A list of completed steps and gathered information essential for achieving the final goal.
- `todo_list` (List[str]): A list of planned future steps; aim to plan multiple steps ahead when possible.
- `experience` (List[str]): Summaries of past experiences and notes, such as failed attempts or special tips, to inform future actions.
- `information` (List[str]): A list of collected important information from previous steps. These records serve as the memory and are important for tasks such as counting (to avoid redundancy).
Here is an example progress state for a task to locate and download a specific paper for analysis:
```python
{
\"completed_list\": [\"Located and downloaded the paper (as 'paper.pdf') using the web agent.\", \"Analyze the paper with the document agent.\"],  # completed steps
\"todo_list\": [\"Perform web search with the key words identified from the paper.\"],  # todo list
\"experience\": [],  # record special notes and tips
\"information\": [\"The required key words from the paper are AI and NLP.\"],  # previous important information
}
```

## Guidelines
1. **Objective**: Update the progress state and adjust plans based on previous outcomes.
2. **Code Generation**: Create a Python dictionary representing the updated state. Ensure it is directly evaluable using the eval function. Check the `Progress State` section above for the required content and format for this dictionary.
3. **Conciseness**: Summarize to maintain a clean and relevant progress state, capturing essential navigation history.
4. **Plan Adjustment**: If previous attempts are unproductive, document insights in the experience field and consider a plan shift. Nevertheless, notice that you should NOT switch plans too frequently.
5. **Utilize Resources**: Effectively employ sub-agents and tools to address sub-tasks.

## Strategies
1. **Be Meticulous and Persistent**:
- Carefully inspect every stage of your process, and re-examine your results if you notice anything unclear or questionable.
- Stay determined -- don't give up easily. If one strategy does not succeed, actively seek out and try different approaches.
2. **Task Decomposition and Execution**:
- **Break Down the Problem**: Divide complex tasks into clear, self-contained sub-tasks. Each sub-task description should include all necessary information, as sub-agents (or tools) do not have access to the full context.
- **Sequential Processing**: Address each sub-task one at a time, typically invoking only one sub-agent (or tool) per step. Review results before proceeding to minimize error propagation.
- **Stable Sub-agent Use**: Treat sub-agents (or tools) as independent helpers. Ensure that each sub-task is well-defined and that input/output types are compatible.
- **Direct LLM Use**: If the remaining problem can be solved by a language model alone (e.g., requires reasoning but no external data), use `ask_llm` to complete the task.
3. **Adaptive Error Handling and Result Integration**:
- **Monitor and Reflect**: After each step, carefully review the outcome -- including any errors, partial results, or unexpected patterns. Use this information to decide whether to retry, switch to an alternative method, or leverage partial results for the next action.
- **Limited Intelligent Retrying**: If the error appears transient or recoverable (e.g., network issues, ambiguous queries), retry the step once (for a total of two attempts). If the error persists after the retry, do not continue; proceed to an alternative method or tool.
- **Alternative Strategies**: If both attempts fail or the error seems fundamental (e.g., tool limitations, unavailable data), switch to an alternative approach to achieve the sub-task's goal.
- **Partial Result Utilization**: Even if a sub-task is not fully completed, examine any partial results or error messages. Use these to inform your next steps; partial data or observed error patterns can guide further actions or suggest new approaches.
- **Leverage Existing Results**: Access results from the Progress State or Recent Steps sections, and use any previously downloaded files in your workspace.
- Avoid writing new code to process results if you can handle them directly.
- Do not assume temporary variables from previous code blocks are still available.
- **Prevent Error Propagation**: By handling one sub-task at a time, reviewing outputs, and adapting based on feedback, you reduce the risk of compounding errors.
4. **Multi-agent Collaboration Patterns**:
- **Step-by-Step Coordination**: When handling complex tasks, coordinate multiple specialized sub-agents (tools) in a step-by-step workflow. To minimize error propagation, use only one sub-agent or tool per step, obtaining its result before proceeding to the next.
- **General Guidelines**:
- **Use sub-agents as modular helpers**: Each sub-agent is already defined and implemented as a function with clearly defined input and output types.
- **Review Definitions**: Carefully review the definitions and documentation strings of each sub-agent and tool in the `Sub-Agent Function` and `Tool Function` sections to understand their use cases. Do not re-define these functions; they are already provided.
- **Explicitly Specify Requirements**: Sub-agents operate independently and do not share context or access external information. Always include all necessary details, instructions, and desired output formats in your queries to each sub-agent.
- **Define Output Formats**: Clearly state the required output format when requesting information to ensure consistency and facilitate downstream processing.
- **Typical Workflows**:
- Example 1, Analyzing a File from the Web: (1) Use `simple_web_search` to find the file\u2019s URL (this step can be optional but might usually be helpful to quickly identify the information source). (2) Use `web_agent` to download the file using the obtained URL (note that web_agent usually cannot access local files). (3) Use `file_agent` to process the downloaded file.
- Example 2, Finding Related Information for a Keyword in a Local File: (1) Use `file_agent` to analyze the file and locate the keyword. (2) Use `simple_web_search` to search for related information. (3) Use `web_agent` to gather more detailed information as needed.
- Complex Tasks: For more complex scenarios, you may need to interleave calls to different sub-agents and tools. Always specify a clear, step-by-step plan.
- **Important Notes**:
- Each sub-agent call is independent; once a call returns, its state is discarded.
- The only channels for sharing information are the input and output of each sub-agent call (and the local file system).
- Maximize the information provided in the input and output to ensure effective communication between steps.

## Sub-Agent Functions
- def web_agent(task: str, target_url: str = None) -> Dict:  # Employs a web browser to navigate and interact with web pages to accomplish a specific task.
- def file_agent(task: str, file_path_dict: dict = None) -> Dict:  # Processes and analyzes one or more files to accomplish a specified task, with support for various file types such as PDF, Excel, and images.

## Tool Functions
- def stop(output: str, log: str) -> Dict:  # Finalize and formalize the answer when the task is complete.
- def ask_llm(query: str) -> str:  # Directly query the language model for tasks that do not require external tools.
- def simple_web_search(query: str) -> str:  # Perform a quick web search using a search engine for straightforward information needs.
"
},
{
"role": "user",
"content": "## Target Task
Please find a single webpage that mentions all of the following information:

On the CMMD dataset, GPT-5 reached a certain percentage abnormality detection.
Performance in mammogram interpretation can vary significantly even among certain people.
Certain models further extend advances in AI for medical imaging.

Your response will be parsed by a program, so make sure to observe the formatting instructions! You need to format your response as follows:
<source>the url of the webpage that you found</source>
...
Make sure to explicitly include `<source>` and `</source>` with surrounding angle brackets in your response, even if you do not have an answer.
If you are unable to find the webpage that mentions all the information, return the following:
<source> No source found. </source>
Make sure to use the `stop` tool to submit your final response.

## Recent Steps


## Previous Progress State
{}

## Target Task (Repeated)
Please find a single webpage that mentions all of the following information:

On the CMMD dataset, GPT-5 reached a certain percentage abnormality detection.
Performance in mammogram interpretation can vary significantly even among certain people.
Certain models further extend advances in AI for medical imaging.

Your response will be parsed by a program, so make sure to observe the formatting instructions! You need to format your response as follows:
<source>the url of the webpage that you found</source>
...
Make sure to explicitly include `<source>` and `</source>` with surrounding angle brackets in your response, even if you do not have an answer.
If you are unable to find the webpage that mentions all the information, return the following:
<source> No source found. </source>
Make sure to use the `stop` tool to submit your final response.

## Output
Please generate your response, your reply should strictly follow the format:
Thought: {Provide an explanation for your planning in one line. Begin with a concise review of the previous steps to provide context. Next, describe any new observations or relevant information obtained since the last step. Finally, clearly explain your reasoning and the rationale behind your current output or decision.}
Code: {Output your python dict of the updated progress state. Remember to wrap the code with \"```python ```\" marks.}
"
}
],
"llm_output": "Thought: Since there are no previous steps or progress state, I will perform a web search to find a webpage that mentions all the required information about the CMMD dataset, GPT-5, mammogram interpretation, and AI advances in medical imaging.

Code: 
```python
{
\"completed_list\": [],
\"todo_list\": [\"Perform a web search to find a webpage that mentions all the required information.\"],
\"experience\": [],
\"information\": []
}
```"
},
"action": {
"thought": "The task requires finding a single webpage that mentions specific information about the CMMD dataset, GPT-5, mammogram interpretation, and AI advances in medical imaging. I will perform a web search to locate a webpage that contains all these details.",
"code": "search_query = \"CMMD dataset GPT-5 abnormality detection mammogram interpretation AI medical imaging\"
result = simple_web_search(search_query)

# Check if the search results contain a webpage with all the required information
if \"CMMD dataset\" in result and \"GPT-5\" in result and \"mammogram interpretation\" in result and \"AI for medical imaging\" in result:
# Extract the URL from the search results
url_start = result.find(\"http\")
url_end = result.find(\" \", url_start)
url = result[url_start:url_end]
output = f\"<source>{url}</source>\"
else:
output = \"<source> No source found. </source>\"

# Use the stop function to finalize the task
print(stop(output=output, log=\"Performed a web search to find a webpage with all required information.\"))",
"observation": {
"output": "<source> No source found. </source>",
"log": "Performed a web search to find a webpage with all required information."
},
"llm_input": [
{
"role": "system",
"content": "You are a strategic assistant responsible for the action module of the Cognitive Kernel, an initial autopilot system designed to accomplish user tasks. Your role is to generate a Python code snippet to execute the next action effectively.

## Available Information
- `Target Task`: The specific task you need to complete.
- `Recent Steps`: The most recent actions you have taken.
- `Progress State`: A JSON representation of the task's progress, including key information and milestones.
- `Sub-Agent Functions` and `Tool Functions`: Definitions of available sub-agents and tools for use in your action code.

## Coding Guidelines
1. **Output Management**: Use Python's built-in `print` function to display results. Printed outputs are used in subsequent steps, so keep them concise and focused on the most relevant information.
2. **Self-Contained Code**: Ensure your code is fully executable without requiring user input. Avoid interactive functions like `input()` to maintain automation and reproducibility.
3. **Utilizing Resources**: Leverage the provided sub-agents and tools, which are essentially Python functions you can call within your code. Notice that these functions are **already defined and imported** and you should NOT re-define or re-import them.
4. **Task Completion**: Use the `stop` function to return a well-formatted output when the task is completed.
5. **Python Environment**: Explicitly import any libraries you need, including standard ones such as `os` or `sys`, as nothing (except for the pre-defined sub-agents and tools) is imported by default. You do NOT have sudo privileges, so avoid any commands or operations requiring elevated permissions.
6. **Working Directory**: Use the current folder as your working directory for reading from or writing to files.
7. **Complexity Control**: Keep your code straightforward and avoid unnecessary complexity, especially when calling tools or sub-agents. Write code that is easy to follow and less prone to errors or exceptions.

## Strategies
1. **Be Meticulous and Persistent**:
- Carefully inspect every stage of your process, and re-examine your results if you notice anything unclear or questionable.
- Stay determined -- don't give up easily. If one strategy does not succeed, actively seek out and try different approaches.
2. **Task Decomposition and Execution**:
- **Break Down the Problem**: Divide complex tasks into clear, self-contained sub-tasks. Each sub-task description should include all necessary information, as sub-agents (or tools) do not have access to the full context.
- **Sequential Processing**: Address each sub-task one at a time, typically invoking only one sub-agent (or tool) per step. Review results before proceeding to minimize error propagation.
- **Stable Sub-agent Use**: Treat sub-agents (or tools) as independent helpers. Ensure that each sub-task is well-defined and that input/output types are compatible.
- **Direct LLM Use**: If the remaining problem can be solved by a language model alone (e.g., requires reasoning but no external data), use `ask_llm` to complete the task.
3. **Adaptive Error Handling and Result Integration**:
- **Monitor and Reflect**: After each step, carefully review the outcome -- including any errors, partial results, or unexpected patterns. Use this information to decide whether to retry, switch to an alternative method, or leverage partial results for the next action.
- **Limited Intelligent Retrying**: If the error appears transient or recoverable (e.g., network issues, ambiguous queries), retry the step once (for a total of two attempts). If the error persists after the retry, do not continue; proceed to an alternative method or tool.
- **Alternative Strategies**: If both attempts fail or the error seems fundamental (e.g., tool limitations, unavailable data), switch to an alternative approach to achieve the sub-task's goal.
- **Partial Result Utilization**: Even if a sub-task is not fully completed, examine any partial results or error messages. Use these to inform your next steps; partial data or observed error patterns can guide further actions or suggest new approaches.
- **Leverage Existing Results**: Access results from the Progress State or Recent Steps sections, and use any previously downloaded files in your workspace.
- Avoid writing new code to process results if you can handle them directly.
- Do not assume temporary variables from previous code blocks are still available.
- **Prevent Error Propagation**: By handling one sub-task at a time, reviewing outputs, and adapting based on feedback, you reduce the risk of compounding errors.
4. **Multi-agent Collaboration Patterns**:
- **Step-by-Step Coordination**: When handling complex tasks, coordinate multiple specialized sub-agents (tools) in a step-by-step workflow. To minimize error propagation, use only one sub-agent or tool per step, obtaining its result before proceeding to the next.
- **General Guidelines**:
- **Use sub-agents as modular helpers**: Each sub-agent is already defined and implemented as a function with clearly defined input and output types.
- **Review Definitions**: Carefully review the definitions and documentation strings of each sub-agent and tool in the `Sub-Agent Function` and `Tool Function` sections to understand their use cases. Do not re-define these functions; they are already provided.
- **Explicitly Specify Requirements**: Sub-agents operate independently and do not share context or access external information. Always include all necessary details, instructions, and desired output formats in your queries to each sub-agent.
- **Define Output Formats**: Clearly state the required output format when requesting information to ensure consistency and facilitate downstream processing.
- **Typical Workflows**:
- Example 1, Analyzing a File from the Web: (1) Use `simple_web_search` to find the file\u2019s URL (this step can be optional but might usually be helpful to quickly identify the information source). (2) Use `web_agent` to download the file using the obtained URL (note that web_agent usually cannot access local files). (3) Use `file_agent` to process the downloaded file.
- Example 2, Finding Related Information for a Keyword in a Local File: (1) Use `file_agent` to analyze the file and locate the keyword. (2) Use `simple_web_search` to search for related information. (3) Use `web_agent` to gather more detailed information as needed.
- Complex Tasks: For more complex scenarios, you may need to interleave calls to different sub-agents and tools. Always specify a clear, step-by-step plan.
- **Important Notes**:
- Each sub-agent call is independent; once a call returns, its state is discarded.
- The only channels for sharing information are the input and output of each sub-agent call (and the local file system).
- Maximize the information provided in the input and output to ensure effective communication between steps.

## Example
### Task:
Summarize a random paper about LLM research from the Web

### Step 1
Thought: Begin by searching the web for recent research papers related to large language models (LLMs).
Code:
```python
search_query = \"latest research paper on large language models\"
result = simple_web_search(search_query)
print(result)
```

### Step 2
Thought: From the search results, choose a random relevant paper. Use web_agent to download the PDF version of the selected paper.
Code:
```python
print(web_agent(task=\"Download the PDF of the arXiv paper 'Large Language Models: A Survey' and save it as './LLM_paper.pdf'\"))
```

### Step 3
Thought: With the paper downloaded, use file_agent to generate a summary of its contents.
Code:
```python
result=file_agent(task=\"Summarize the paper\", file_path_dict={\"./LLM_paper.pdf\": \"Large Language Models: A Survey\"})
print(result)
```

### Note
- Each step should be executed sequentially, generating and running the code for one step at a time.
- Ensure that the action codes for each step are produced and executed independently, not all at once.

## Sub-Agent Functions
- web_agent
```python
def web_agent(task: str) -> dict:
\"\"\" Employs a web browser to navigate and interact with web pages to accomplish a specific task.
Args:
task (str): A detailed description of the task to perform. This may include:
- The target website(s) to visit (include valid URLs).
- Specific output formatting requirements.
- Instructions to download files (specify desired output path if needed).
Returns:
dict: A dictionary with the following structure:
{
'output': <str>  # The well-formatted answer, strictly following any specified output format.
'log': <str> # Additional notes, such as steps taken, issues encountered, or relevant context.
}
Notes:
- If the `task` specifies an output format, ensure the 'output' field matches it exactly.
- The web agent can download files, but cannot process or analyze them. If file analysis is required, save the file to a local path and return control to an external planner or file agent for further processing.
Example:
>>> answer = web_agent(task=\"What is the current club of Messi? (Format your output directly as 'club_name'.)\")
>>> print(answer)  # directly print the full result dictionary
\"\"\"
```
- file_agent
```python
def file_agent(task: str, file_path_dict: dict = None) -> dict:
\"\"\" Processes and analyzes one or more files to accomplish a specified task.
Args:
task (str): A clear description of the task to be completed. If the task requires a specific output format, specify it here.
file_path_dict (dict, optional): A dictionary mapping file paths to short descriptions of each file.
Example: {\"./data/report.pdf\": \"Annual financial report for 2023.\"}
If not provided, file information may be inferred from the task description.
Returns:
dict: A dictionary with the following structure:
{
'output': <str>  # The well-formatted answer to the task.
'log': <str> # Additional notes, processing details, or error messages.
}
Notes:
- If the task specifies an output format, ensure the `output` field matches that format.
- Supports a variety of file types, including but not limited to PDF, Excel, images, etc.
- If no files are provided or if files need to be downloaded from the Internet, return control to the external planner to invoke a web agent first.
Example:
>>> answer = file_agent(task=\"Based on the files, what was the increase in total revenue from 2022 to 2023?? (Format your output as 'increase_percentage'.)\", file_path_dict={\"./downloadedFiles/revenue.pdf\": \"The financial report of the company XX.\"})
>>> print(answer)  # directly print the full result dictionary
\"\"\"
```

## Tool Functions
- stop
```python
def stop(output: str, log: str) -> dict:
\"\"\" Finalize and formalize the answer when the task is complete.
Args:
output (str): The concise, well-formatted final answer to the task.
log (str): Brief notes or reasoning about how the answer was determined.
Returns:
dict: A dictionary with the following structure:
{
'output': <str>  # The well-formatted answer, strictly following any specified output format.
'log': <str> # Additional notes, such as steps taken, issues encountered, or relevant context.
}
Examples:
>>> answer = stop(output=\"Inter Miami\", log=\"Task completed. The answer was found using official team sources.\")
>>> print(answer)
\"\"\"
```
- ask_llm
```python
def ask_llm(query: str) -> str:
\"\"\" Directly query the language model for tasks that do not require external tools.
Args:
query (str): The specific question or instruction for the LLM.
Returns:
str: The LLM's generated response.
Notes:
- Use this function for fact-based or reasoning tasks that can be answered without web search or external data.
- Phrase the query clearly and specifically.
Examples:
>>> answer = ask_llm(query=\"What is the capital city of the USA?\")
>>> print(answer)
\"\"\"
```
- simple_web_search
```python
def simple_web_search(query: str) -> str:
\"\"\" Perform a quick web search using a search engine for straightforward information needs.
Args:
query (str): A simple, well-phrased search term or question.
Returns:
str: A string containing search results, including titles, URLs, and snippets.
Notes:
- Use for quick lookups or when you need up-to-date information.
- Avoid complex or multi-step queries; keep the query simple and direct.
- Do not use for tasks requiring deep reasoning or multi-source synthesis.
Examples:
>>> answer = simple_web_search(query=\"latest iPhone\")
>>> print(answer)
\"\"\"
```
"
},
{
"role": "user",
"content": "## Target Task
Please find a single webpage that mentions all of the following information:

On the CMMD dataset, GPT-5 reached a certain percentage abnormality detection.
Performance in mammogram interpretation can vary significantly even among certain people.
Certain models further extend advances in AI for medical imaging.

Your response will be parsed by a program, so make sure to observe the formatting instructions! You need to format your response as follows:
<source>the url of the webpage that you found</source>
...
Make sure to explicitly include `<source>` and `</source>` with surrounding angle brackets in your response, even if you do not have an answer.
If you are unable to find the webpage that mentions all the information, return the following:
<source> No source found. </source>
Make sure to use the `stop` tool to submit your final response.

## Recent Steps


## Progress State
{
  \"completed_list\": [],
  \"todo_list\": [
\"Perform a web search to find a webpage that mentions all the required information.\"
  ],
  \"experience\": [],
  \"information\": []
}

## Target Task (Repeated)
Please find a single webpage that mentions all of the following information:

On the CMMD dataset, GPT-5 reached a certain percentage abnormality detection.
Performance in mammogram interpretation can vary significantly even among certain people.
Certain models further extend advances in AI for medical imaging.

Your response will be parsed by a program, so make sure to observe the formatting instructions! You need to format your response as follows:
<source>the url of the webpage that you found</source>
...
Make sure to explicitly include `<source>` and `</source>` with surrounding angle brackets in your response, even if you do not have an answer.
If you are unable to find the webpage that mentions all the information, return the following:
<source> No source found. </source>
Make sure to use the `stop` tool to submit your final response.

## Output
Please generate your response, your reply should strictly follow the format:
Thought: {Provide an explanation for your action in one line. Begin with a concise review of the previous steps to provide context. Next, describe any new observations or relevant information obtained since the last step. Finally, clearly explain your reasoning and the rationale behind your current output or decision.}
Code: {Output your python code blob for the next action to execute. Remember to wrap the code with \"```python ```\" marks and `print` your output.}
"
}
],
"llm_output": "Thought: The task requires finding a single webpage that mentions specific information about the CMMD dataset, GPT-5, mammogram interpretation, and AI advances in medical imaging. I will perform a web search to locate a webpage that contains all these details.

Code:
```python
search_query = \"CMMD dataset GPT-5 abnormality detection mammogram interpretation AI medical imaging\"
result = simple_web_search(search_query)

# Check if the search results contain a webpage with all the required information
if \"CMMD dataset\" in result and \"GPT-5\" in result and \"mammogram interpretation\" in result and \"AI for medical imaging\" in result:
# Extract the URL from the search results
url_start = result.find(\"http\")
url_end = result.find(\" \", url_start)
url = result[url_start:url_end]
output = f\"<source>{url}</source>\"
else:
output = \"<source> No source found. </source>\"

# Use the stop function to finalize the task
print(stop(output=output, log=\"Performed a web search to find a webpage with all required information.\"))
```"
},
"end": {
"thought": "The task is completed, but no webpage was found that mentions all the required information. Therefore, I will return the specified output for when no source is found.

```python
{
\"output\": \"<source> No source found. </source>\",
\"log\": \"Performed a web search to find a webpage with all required information.\"
}
```",
"code": "{
\"output\": \"<source> No source found. </source>\",
\"log\": \"Performed a web search to find a webpage with all required information.\"
}",
"llm_input": [
{
"role": "system",
"content": "You are a proficient assistant tasked with generating a well-formatted output for the execution of a specific task by an agent.

## Available Information
- `Target Task`: The specific task to be accomplished.
- `Recent Steps`: The latest actions taken by the agent.
- `Progress State`: A JSON representation of the task's progress, detailing key information and advancements.
- `Final Step`: The last action before the agent's execution concludes.
- `Stop Reason`: The reason for stopping. If the task is considered complete, this will be \"Normal Ending\".
- `Result of Direct ask_llm` (Optional): For the case where the task is likely to be incomplete, we have an alternative response by directly asking a stand-alone LLM.

## Guidelines
1. **Goal**: Deliver a well-formatted output. Adhere to any specific format if outlined in the task instructions.
2. **Code**: Generate a Python dictionary representing the final output. It should include two fields: `output` and `log`. The `output` field should contain the well-formatted final output result, while the `log` field should summarize the navigation trajectory.
3. **Final Result**: Carefully examine the outputs from the previous steps as well as the alternative result (if existing) to decide the final output.
4. **Output Rules**: Your final output should be a number OR as few words as possible OR a comma separated list of numbers and/or strings. Do NOT include any unnecessary information in the output.
- **Number**: If you are asked for a number, directly output the number itself. Don't use comma to write your number. Be careful about what the question is asking, for example, the query might ask \"how many thousands\", in this case, you should properly convert the number if needed. Nevertheless, do NOT include the units (like $, %, km, thousands and so on) unless specified otherwise.
- **String**: If you are asked for a string, don't use articles, neither abbreviations (e.g. for cities), and write the digits in plain text unless specified otherwise.
- **List**: If you are asked for a comma separated list, apply the above rules depending of whether the element to be put in the list is a number or a string.

## Examples
Here are some example outputs:

Thought: The task is completed with the requested price found and I should directly output the price.
Code:
```python
{
\"output\": \"799\",  # provide a well-formatted output
\"log\": \"The task is completed. The result is found by first using the web_agent to obtain the information and then using Python for calculation.\",  # a summary of the navigation details
}
```

Thought: The task is incomplete with the problem of exceeding max steps, and I choose to trust the results of direct ask_llm.
Code:
```python
{
\"output\": \"799\",
\"log\": \"The alternative result by directly asking an LLM is adopted since our main problem-solving procedure was incomplete.\",
}
```
"
},
{
"role": "user",
"content": "## Target Task
Please find a single webpage that mentions all of the following information:

On the CMMD dataset, GPT-5 reached a certain percentage abnormality detection.
Performance in mammogram interpretation can vary significantly even among certain people.
Certain models further extend advances in AI for medical imaging.

Your response will be parsed by a program, so make sure to observe the formatting instructions! You need to format your response as follows:
<source>the url of the webpage that you found</source>
...
Make sure to explicitly include `<source>` and `</source>` with surrounding angle brackets in your response, even if you do not have an answer.
If you are unable to find the webpage that mentions all the information, return the following:
<source> No source found. </source>
Make sure to use the `stop` tool to submit your final response.

## Recent Steps


## Progress State
{
  \"completed_list\": [],
  \"todo_list\": [
\"Perform a web search to find a webpage that mentions all the required information.\"
  ],
  \"experience\": [],
  \"information\": []
}

## Final Step
Thought: The task requires finding a single webpage that mentions specific information about the CMMD dataset, GPT-5, mammogram interpretation, and AI advances in medical imaging. I will perform a web search to locate a webpage that contains all these details.
Action: ```
search_query = \"CMMD dataset GPT-5 abnormality detection mammogram interpretation AI medical imaging\"
result = simple_web_search(search_query)

# Check if the search results contain a webpage with all the required information
if \"CMMD dataset\" in result and \"GPT-5\" in result and \"mammogram interpretation\" in result and \"AI for medical imaging\" in result:
# Extract the URL from the search results
url_start = result.find(\"http\")
url_end = result.find(\" \", url_start)
url = result[url_start:url_end]
output = f\"<source>{url}</source>\"
else:
output = \"<source> No source found. </source>\"

# Use the stop function to finalize the task
print(stop(output=output, log=\"Performed a web search to find a webpage with all required information.\"))```
Observation: {'output': '<source> No source found. </source>', 'log': 'Performed a web search to find a webpage with all required information.'}

## Stop Reason
Normal Ending. (with the result of {'output': '<source> No source found. </source>', 'log': 'Performed a web search to find a webpage with all required information.'})

## Target Task (Repeated)
Please find a single webpage that mentions all of the following information:

On the CMMD dataset, GPT-5 reached a certain percentage abnormality detection.
Performance in mammogram interpretation can vary significantly even among certain people.
Certain models further extend advances in AI for medical imaging.

Your response will be parsed by a program, so make sure to observe the formatting instructions! You need to format your response as follows:
<source>the url of the webpage that you found</source>
...
Make sure to explicitly include `<source>` and `</source>` with surrounding angle brackets in your response, even if you do not have an answer.
If you are unable to find the webpage that mentions all the information, return the following:
<source> No source found. </source>
Make sure to use the `stop` tool to submit your final response.

## Output
Please generate your response, your reply should strictly follow the format:
Thought: {First, within one line, explain your reasoning for your outputs. Carefully review the output format requirements from the original task instructions (`Target Task`) and the rules from the `Output Rules` section to ensure your final output meets all specifications.}
Code: {Then, output your python dict of the final output. Remember to wrap the code with \"```python ```\" marks.}
"
}
],
"llm_output": "Thought: The task is completed, but no webpage was found that mentions all the required information. Therefore, I will return the specified output for when no source is found.

```python
{
\"output\": \"<source> No source found. </source>\",
\"log\": \"Performed a web search to find a webpage with all required information.\"
}
```",
"final_results": {
"output": "<source> No source found. </source>",
"log": "Performed a web search to find a webpage with all required information."
}
}
}
]
},
"eval": {
"pred": "<source> No source found. </source>",
"gold": "None",
"corr": 0
},
"source": "No source found."
\end{lstlisting}

\section{System prompts used in query generation and evaluation}

\subsection{Claim Extraction}

\begin{lstlisting}[
    frame=single
]
You need to extract all the claims from the given article, formulating them as a list of declarative sentences. The claims should be self-contained, so you must avoid using pronouns or relative time references. Only focus on the contents of the article, and ignore the source, author, contributor, or any other information that is not part of the article itself. Only include claims that are clear, factual and verifiable. Do not include anything that is based on your interpretation.
\end{lstlisting}

\subsection{Central Element Masking}

\begin{lstlisting}[
    frame=single
]
You will be given an article and a list of claims extracted from it. For each of the claims, you need to mask the central part of it, replacing the central part of it with a generic expression. For each claim, only mask ONE element of it. For different kinds of information you need to mask, you may use `someone` to replace a person's name, `something` to replace a certain thing, `in a certain way` to replace a certain action or process, `in a certain state` to replace some adjectives, etc. Importantly, whenever a piece of information is masked, it should not appear in any of the other masked claims.
\end{lstlisting}

\subsection{Quey Template}

\begin{lstlisting}[
    frame=single
]
Please find a single webpage that mentions all of the following information:

{question}

Your response will be parsed by a program, so make sure to observe the formatting instructions! You need to format your response as follows:
<source>the url of the webpage that you found</source>
...
Make sure to explicitly include `<source>` and `</source>` with surrounding angle brackets in your response, even if you do not have an answer.
If you are unable to find the webpage that mentions all the information, return the following:
<source> No source found. </source>
\end{lstlisting}

\subsection{Source Checking}

This prompt is used to check whether a masked claim is mentioned in the source.

\begin{lstlisting}[
    frame=single
]
You are an expert at extracting information from webpages. You will be given a piece of information, and the content of the webpage that is cited as the source. Your task is to determine whether the information is explicitly mentioned in the contents of the webpage.
Your response will be parsed by a program, so make sure to observe the formatting instructions! Format your response as follows, if the information is explicitly mentioned in the contents:
<accept> The reason why the information is mentioned in the contents. </accept>
If the information is NOT explicitly mentioned in the contents, return:
<reject> The reason why the information is NOT mentioned in the contents. </reject>
Make sure to explicitly include `<accept>` and `</accept>`, or `<reject>` and `</reject>` with surrounding angle brackets in your response.
\end{lstlisting}

\subsection{Exact Source Checking}

This prompt is used to check whether an original, unmasked claim is explicitly mentioned in the source.

\begin{lstlisting}[
    frame=single
]
You are an expert at extracting information from webpages. You will be given a claim, and the content of the webpage that is cited as the source. Your task is to determine whether the claim is explicitly mentioned in the contents of the webpage.
Your response will be parsed by a program, so make sure to observe the formatting instructions! Format your response as follows, if the claim is explicitly mentioned in the contents:
<accept> The reason why the claim is mentioned in the contents. </accept>
If the claim is NOT explicitly mentioned in the contents, return:
<reject> The reason why the claim is NOT mentioned in the contents. </reject>
Make sure to explicitly include `<accept>` and `</accept>`, or `<reject>` and `</reject>` with surrounding angle brackets in your response.
\end{lstlisting}

\end{document}